\documentclass{article}

\usepackage[preprint]{neurips_2026}
\usepackage[table]{xcolor}
\usepackage[pdftex]{graphicx}
\usepackage{amssymb}
\usepackage{pifont}
\usepackage{booktabs}
\usepackage{colortbl}
\usepackage[utf8]{inputenc}
\usepackage[T1]{fontenc}
\usepackage{hyperref}
\usepackage{url}
\usepackage{amsfonts}
\usepackage{amsmath}
\usepackage{graphicx}
\usepackage{enumitem}
\usepackage{tikz}
\usepackage[most]{tcolorbox}
\usepackage{xcolor}
\usepackage{listings}
\usepackage{caption}
\usepackage{subcaption}
\usepackage{fontawesome5}
\usepackage{tabularx}

\usetikzlibrary{positioning,fit,calc,shadows.blur,backgrounds,arrows.meta}
\usetikzlibrary{shapes.geometric}

\definecolor{paperbg}{RGB}{248,248,246}
\definecolor{codebg}{RGB}{245,247,250}
\definecolor{framecol}{RGB}{210,214,220}
\definecolor{paperbg}{RGB}{248,248,246}
\definecolor{framecol}{RGB}{210,214,220}
\definecolor{paperbg}{RGB}{248,248,246}
\definecolor{framecol}{RGB}{210,214,220}

\definecolor{ghbg}{RGB}{244,250,244}
\definecolor{ghborder}{RGB}{208,215,222}
\definecolor{ghkeyword}{RGB}{17,99,41}\definecolor{ghstring}{RGB}{10,48,105}
\definecolor{ghcomment}{RGB}{87,96,106}
\definecolor{ghnumber}{RGB}{111,66,193}
\definecolor{ghname}{RGB}{36,41,47}
\definecolor{ghlinenumber}{RGB}{139,148,158}
\lstdefinestyle{githubpython}{
    language=Python,
    backgroundcolor=\color{ghbg},
    escapeinside={(*@}{@*)},
    basicstyle=\ttfamily\scriptsize\color{ghname},
    keywordstyle=\color{ghkeyword}\bfseries,
    stringstyle=\color{ghstring},
    commentstyle=\color{ghcomment}\itshape,
    numberstyle=\tiny\color{ghlinenumber},
    numbers=left,
    numbersep=6pt,
    xleftmargin=10pt,
    frame=single,
    rulecolor=\color{ghborder},
    framerule=0.4pt,
    columns=fullflexible,
    keepspaces=true,
    showstringspaces=false,
    tabsize=4,
    breaklines=true,
    aboveskip=0pt,
    belowskip=0pt
}

\definecolor{taskbg}{RGB}{248,250,252}
\definecolor{taskframe}{RGB}{190,198,210}
\definecolor{tasktitle}{RGB}{35,55,90}
\definecolor{agentbg}{RGB}{240,250,243}
\definecolor{agentframe}{RGB}{115,170,125}
\definecolor{agenttitle}{RGB}{35,105,60}

\newtcolorbox{agentcard}[1][]{
    enhanced,
    breakable,
    colback=agentbg,
    colframe=agentframe,
    coltitle=white,
    colbacktitle=agenttitle,
    title=#1,
    fonttitle=\bfseries,
    boxrule=0.6pt,
    arc=2mm,
    left=2mm,
    right=2mm,
    top=1.5mm,
    bottom=1.5mm,
    before skip=1em,
    after skip=1em
}

\newtcolorbox{taskcard}[1][]{
    enhanced,
    breakable,
    colback=taskbg,
    colframe=taskframe,
    coltitle=white,
    colbacktitle=tasktitle,
    title=#1,
    fonttitle=\bfseries,
    boxrule=0.6pt,
    arc=2mm,
    left=2mm,
    right=2mm,
    top=1.5mm,
    bottom=1.5mm,
    before skip=1em,
    after skip=1em
}

\definecolor{judgeboxbg}{RGB}{248,250,252}
\definecolor{judgeframe}{RGB}{190,198,210}
\definecolor{judgetitle}{RGB}{45,62,90}
\definecolor{jsonbg}{RGB}{246,248,250}

\newtcolorbox{judgebox}[1][]{
    enhanced,
    breakable,
    colback=judgeboxbg,
    colframe=judgeframe,
    coltitle=white,
    colbacktitle=judgetitle,
    title=#1,
    fonttitle=\bfseries,
    boxrule=0.6pt,
    arc=2mm,
    left=2mm,
    right=2mm,
    top=1.5mm,
    bottom=1.5mm,
    before skip=1em,
    after skip=1em
}

\lstdefinestyle{jsonstyle}{
    basicstyle=\ttfamily\scriptsize,
    backgroundcolor=\color{jsonbg},
    frame=single,
    rulecolor=\color{judgeframe},
    breaklines=true,
    showstringspaces=false,
    columns=fullflexible,
    keepspaces=true
}

\definecolor{taskbg}{RGB}{248,250,252}
\definecolor{taskframe}{RGB}{190,198,210}
\definecolor{tasktitle}{RGB}{45,62,90}
\definecolor{codebg}{RGB}{246,248,250}

\newtcolorbox{filecard}[4]{%
    enhanced,
    breakable,
    colback=#1,
    colframe=#2,
    coltitle=white,
    colbacktitle=#3,
    title=#4,
    fonttitle=\bfseries,
    boxrule=0.6pt,
    arc=2mm,
    left=2mm,
    right=2mm,
    top=1.5mm,
    bottom=1.5mm,
    before skip=1em,
    after skip=1em
}

\lstdefinestyle{yamlplain}{
    basicstyle=\ttfamily\scriptsize,
    backgroundcolor=\color{codebg},
    frame=single,
    rulecolor=\color{taskframe},
    breaklines=true,
    showstringspaces=false,
    columns=fullflexible,
    keepspaces=true
}

\newcommand{\ColliderBench}{\textsc{Collider-Bench}}

\title{\ColliderBench{}:\\ Benchmarking AI Agents with Particle Physics Analysis Reproduction}

\author{
  Darius A. Faroughy \\
  New High Energy Theory Center \\
  Department of Physics \& Astronomy\\
  Rutgers University\\
  \texttt{darius.faroughy@rutgers.edu} \\
  \And
  Sofia Palacios Schweitzer \\
  New High Energy Theory Center \\
  Department of Physics \& Astronomy\\
  Rutgers University \\
  \texttt{spalacios@physics.rutgers.edu} \\
  \And
  Ian Pang \\
  New High Energy Theory Center \\
  Department of Physics \& Astronomy\\
  Rutgers University \\
  \texttt{ian.pang@physics.rutgers.edu} \\
  \And
  Siddharth Mishra-Sharma \\
  Faculty of Computing \& Data Sciences \\ 
  Boston University \\
  \texttt{smishras@bu.edu} \\
  \And
  David Shih \\
  New High Energy Theory Center \\
  Department of Physics \& Astronomy\\
  Rutgers University \\
  \texttt{shih@physics.rutgers.edu} \\
}

\begin{document}
\maketitle

\begin{abstract}
Autonomous language-model agents are increasingly evaluated on long-horizon tool-use tasks, but existing benchmarks rarely capture the complexity and nuance of real scientific work. To address this gap, we introduce \ColliderBench{}, a benchmark for evaluating whether LLM agents can reproduce experimental analyses from the Large Hadron Collider (LHC) using only public papers and open scientific software. Such analyses are often difficult to reproduce because the public toolchain only approximates the software used internally by the experimental collaborations, while the published papers inevitably omit implementation details needed for a faithful reconstruction. Agents must therefore rely on physical reasoning, domain knowledge, and trial-and-error to fill these gaps. Each task requires the agent to turn a published analysis into an executable simulation-and-selection pipeline and submit predicted collision event yields in specified signal regions. These predictions are evaluated with standard histogram metrics that provide continuous fidelity scores without a hand-written rubric. We also report the computational cost incurred by each agent per task. Finally, we evaluate the codebase and full session trace using an LLM judge to catch qualitative failure modes such as fabrications, hallucinations and duplications. We release an initial set of tasks drawn from LHC searches, together with a† containerized sandbox and event simulation tools. We evaluate across a capability ladder of general purpose coding agents. Our results show that on average no agent reliably beats the physicist-in-the-loop solution.\\ 
\end{abstract}

\section{Introduction}
\label{sec:intro}

Large language model agents are now capable of taking on substantial parts of scientific workflows, raising the question of how to evaluate them rigorously. Although much of scientific work is mechanical once the configuration is set, the choices that fix that configuration, i.e. which inputs to use, which approximations are defensible, how to reconcile inconsistencies in the source material, depend on reasoning and judgment rather than mere execution. A growing set of science-agent benchmarks evaluate agents on individual analysis steps, reproduction from authored code, or end-to-end paper reproduction scored against expert-authored rubrics.

We introduce \ColliderBench{}, a new benchmark for evaluating  autonomous LLM agents on long-horizon, real-world scientific tasks in a novel setting -- analyses from the Large Hadron Collider (LHC). The agent must perform an end-to-end reproduction of a published analysis against a quantitative target. This must be achieved despite missing or underspecified information -- often held internally by experimental collaborations -- and a public toolchain that only approximates the original experimental setup.
 
In each \ColliderBench{} task, the agent receives a structured prompt specifying the prediction to reproduce, the target publication, a fixed set of CLI tools wrapping the public simulation software, and a containerized sandbox that isolates the run from other tasks. It must read the publication to fill in the details the brief omits and locate the relevant public inputs. It then simulates events of a specified new-physics model through a multi-tool generation and detector-response pipeline, and runs the simulated events through an analysis script implementing the publication's selection logic. Finally, it must produce a binned histogram of predicted counts together with the code and configurations that produced it. For the first release of \ColliderBench{}, we release 10 initial tasks drawn from four LHC analyses. Many more tasks and analyses (at least an order of magnitude more) are available and will be added in future releases. The four source papers span a range of difficulty, varying in whether their analyses use standard or non-standard event-selection features and in how sensitive the predicted yields are to simulation choices that the publication does not fully specify. Each task was first solved by a domain expert working with an agent through the same toolchain, establishing that the public stack can reproduce the published target.

We evaluate each submission with physics-motivated quantitative metrics. Each agent is run autonomously, in a single session without human intervention, using off-the-shelf vendor CLIs and the model's maximum reasoning effort, with a 2.5-hour wall-clock budget. To control for run-to-run stochasticity in agent behavior, every agent is evaluated three times on each task. Because the published reference values are themselves available in the source publications or in auxiliary material, we complement the quantitative score with an LLM judge  \cite{zheng2023judging}  that inspects the agent's full workspace and verifies that the submitted values trace to an executed simulation, not to fabricated scaling factors or values copied from the literature. We report results across frontier models from~\citep{anthropic2026claudemodels}, ~\citep{ openai2026gpt54thinking,openai2026gpt55}, and~\citep{deepseek2026v4}, and additionally probe capability ladders within the Anthropic and OpenAI families to study how performance scales with model capability.

\ColliderBench{} ships the task corpus with hidden reference values, a containerized execution environment, a thin agent harness compatible with multiple frontier agent CLIs, an LLM provenance judge, and a single-session baseline agent. Code, container image, and tasks are released here: 

{
\small
 \includegraphics[height=0.95em]{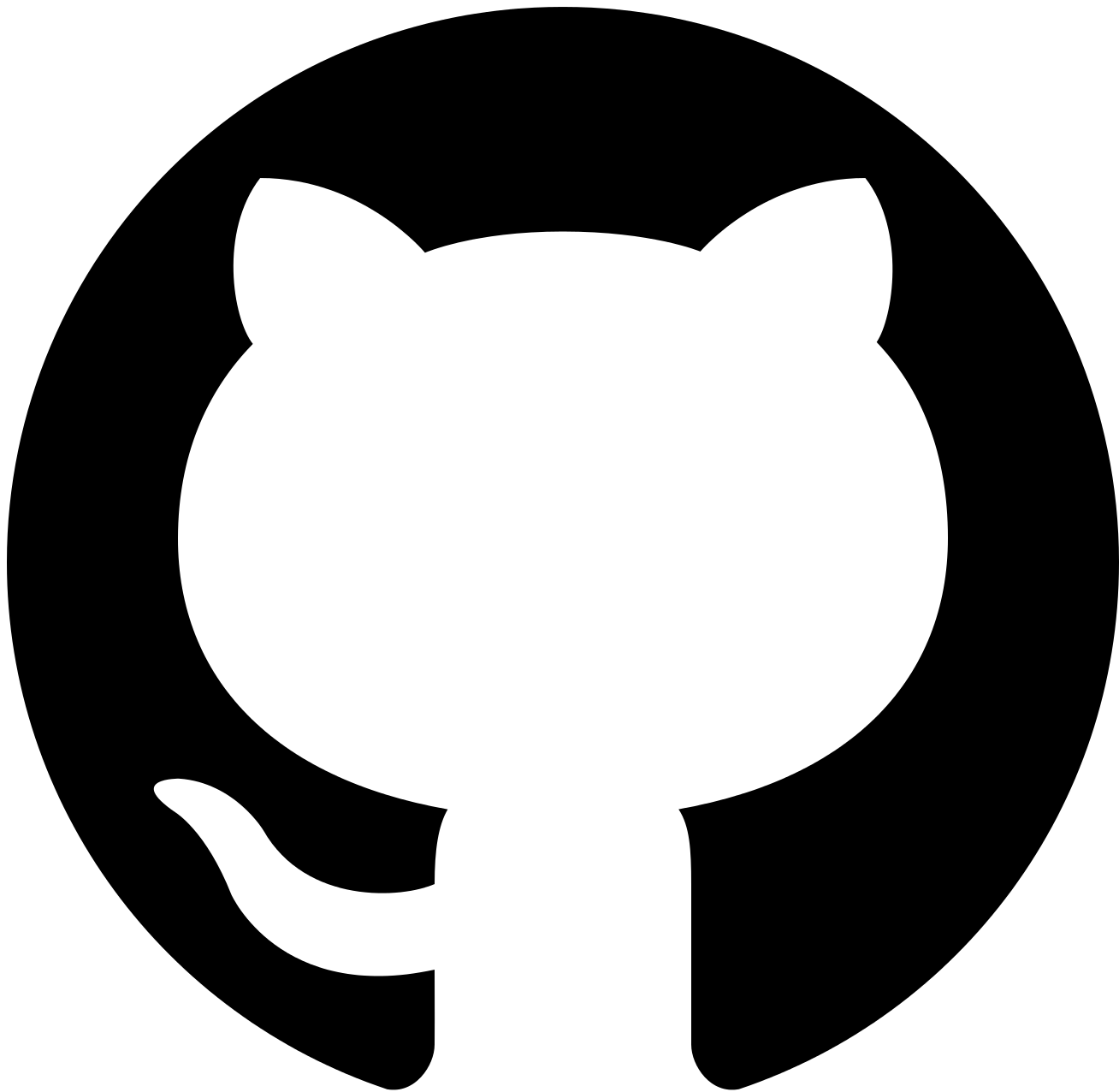} ~\href{https://github.com/dfaroughy/Collider-Bench}{\tt https://github.com/dfaroughy/Collider-Bench}.\\
\includegraphics[height=0.95em]{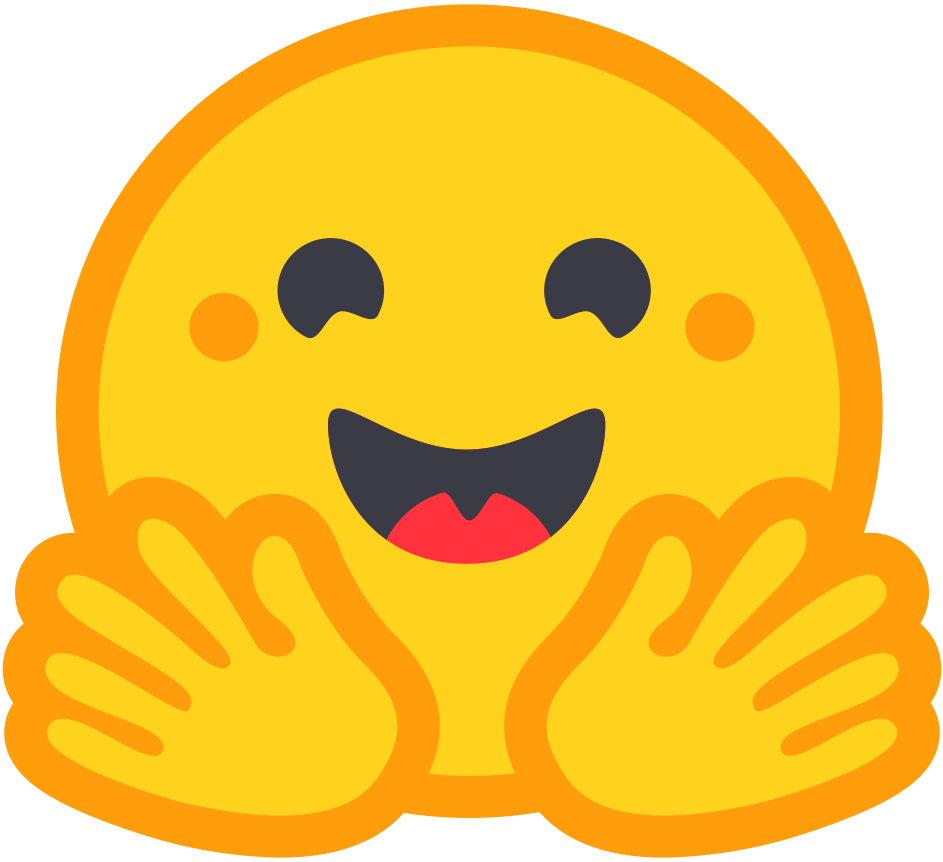}
~\href{https://huggingface.co/datasets/Dariusfar/ColliderBench}{\tt https://huggingface.co/datasets/Dariusfar/ColliderBench}
}

\section{Related Work}
\label{sec:related}

\paragraph{LLM Benchmarks for Science}
A growing set of benchmarks evaluates LLMs on scientific knowledge and reasoning through self-contained question-answer problems. ChemBench~\citep{mirza2025chembench} and LAB-Bench~\citep{labbench2024} probe chemistry and biology knowledge through curated question banks scored against expert-curated answers. FrontierScience~\citep{frontierscience2025} extends this to Olympiad-style and PhD-level research questions in physics, chemistry, and biology, scored either by exact match on the final answer or by an expert-designed rubric. CURIE~\citep{cui2025curie} tests long-context understanding across scientific documents. SciCode~\citep{tian2024scicode} evaluates scientific code generation against scientist-curated test cases, and TPBench~\citep{Chung:2025nsd} evaluates theoretical physics derivations with auto-checked answers. These benchmarks capture LLMs' scientific reasoning at the level of a self-contained problem, but unlike \ColliderBench{} they do not test whether models can construct and execute the multi-step computational pipelines that usually produce real scientific results.

\paragraph{Agent Benchmarks for multi-step Workflows}
A second class of benchmarks evaluates agents, namely LLMs equipped with tools and an execution environment, on multi-step workflows. General-purpose agent benchmarks evaluate code modification, terminal use, and end-to-end ML engineering~\citep{jimenez2024swebench, terminalbench2025, chan2024mlebench, wijk2024rebench}. Scientific benchmarks so far are evaluating isolated analysis steps~\citep{chen2024scienceagentbench}, reproduction from authored code~\citep{siegel2024corebench}, replication of ML-research scored against expert-authored rubrics~\citep{openai2025paperbench}, end-to-end replication of astrophysics papers~\citep{ye2025replicationbench}, biomedical ML workflows~\citep{bioml2025}, or law discovery in physics simulators~\citep{newtonbench2025, gravitybench2025}. \ColliderBench{} targets autonomous pipeline construction across multiple new-physics analyses, with quantitative comparison against published targets, a combination not addressed by any of the existing benchmarks.

\paragraph{Recasting and AI Agents in Particle Physics}
Reproducing published experimental analyses is a long-standing practice in particle physics, formalised in the recasting program~\citep{LHCReinterpretationForum:2020xtr}. Several recent works demonstrate LLM agents on adjacent simulation and analysis tasks~\citep{Gendreau-Distler:2025fsj, Menzo:2025cim, Moreno:2026mqk, Plehn:2026gxv, Qiu:2026iby, Agrawal:2026lvg}. These contribute scaffolding designs and proof-of-principle demonstrations, but they are not benchmarks. They assess execution success rather than correctness against a quantitative reference, and do not evaluate whether their scaffolding adds value over running off-the-shelf frontier agents directly on the same task. \ColliderBench{} fills this gap by providing the quantitative evaluation infrastructure to make such comparisons possible, on a curated set of real published analyses and without human intervention in the run.

\section{\ColliderBench{} }
\label{sec:benchmark}

\begin{figure}[t]
    \centering
    \includegraphics[width=1\textwidth]{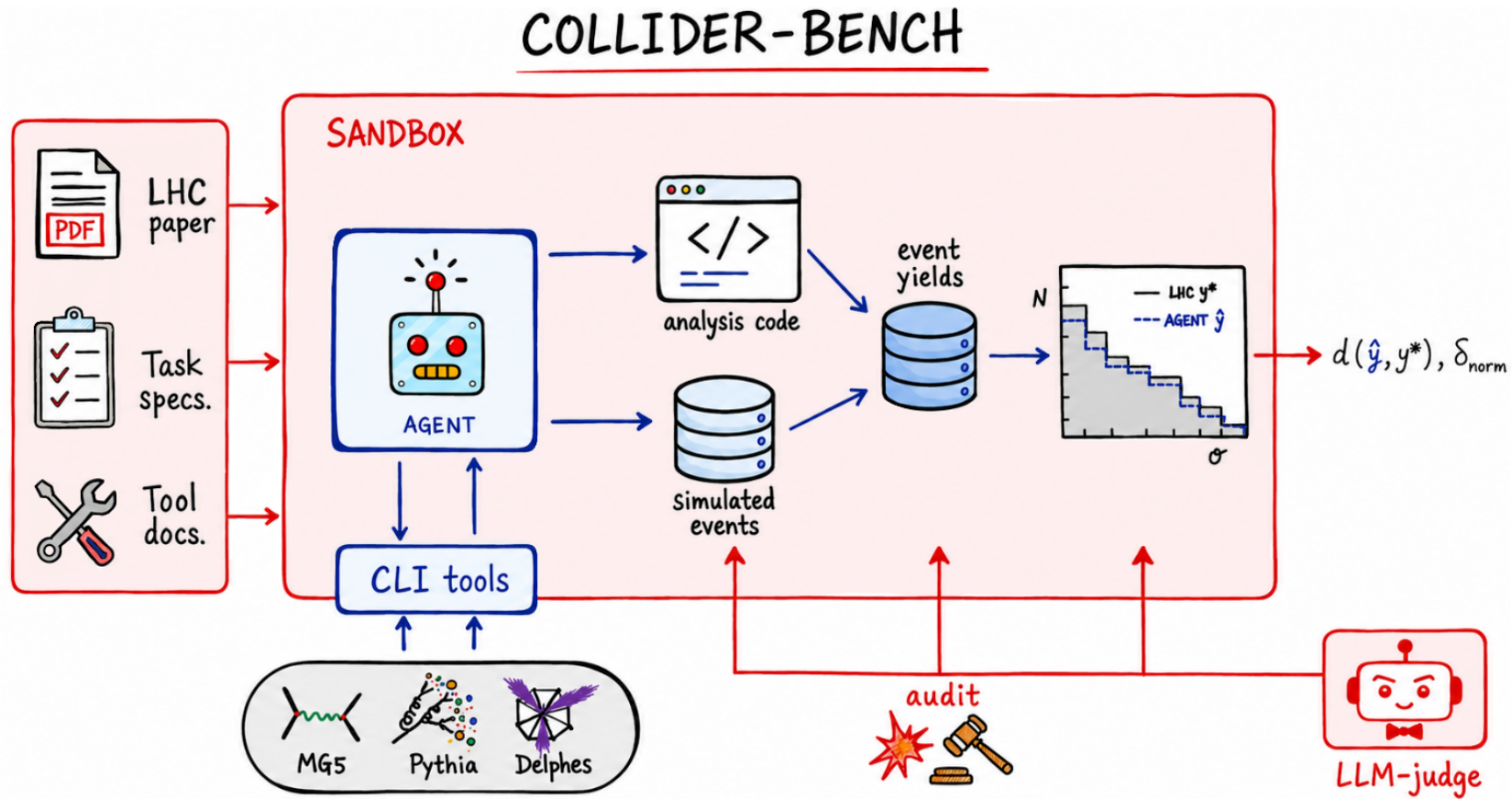}
    
\caption{{Overview of the \ColliderBench{} workflow. The agent writes analysis code and generates events via the CLI tools, which interface with the public simulation stack. Yields are aggregated into a binned histogram and scored against the published reference. An LLM judge, outside the sandbox, audits the agent's workspace.} 
}
\label{fig:overview}
\end{figure}

A schematic of the benchmark architecture, including the paper, output template, tools, sandbox, agent, and evaluator, is shown in Figure~\ref{fig:overview}. Appendix~\ref{app:collider-primer} provides a brief introduction to LHC physics and the simulation tools together with a glossary defining the domain-specific terminology used throughout the paper.

\subsection{Problem Formulation}
\label{sec:problem}

A collider analysis 
can be viewed as a statistical query over collision events.
The collider analyses curated by
\ColliderBench{} are {\it searches for new physics}. These are searches for hypothetical new elementary particles and interactions beyond the Standard Model of particle physics. The hypothetical new physics forms the basis for the ``signal'' and the ordinary physics predicted by the Standard Model is the ``background''. Each search is a stringent statistical test of the validity of the Standard Model, and discovery of new physics in any search (ruling out the null, background-only hypothesis of the Standard Model) would be a paradigm-shifting event worthy of a Nobel Prize in Physics.

The vast majority of searches at the LHC are event-counting experiments. Starting from reconstructed event records, an analysis defines physical objects such as photons, electrons, muons, and collimated jets of hadrons, then applies kinematic selection criteria to isolate events of interest. The surviving events are grouped into {\it signal regions}, where the hypothetical new physics is expected to be enriched over the background. Each analysis reports event yields or binned distributions in signal regions used to test hypotheses for new physics. 

\emph{Recasting} is the process of reusing a published search to constrain signal models different from those explicitly considered in the original analysis. Because researchers outside the LHC experimental collaborations typically lack access to the full detector reconstruction, official signal simulations, and internal analysis code, a recast must approximate the analysis using public information. This typically involves generating signal events with standard Monte Carlo tools, applying a fast detector simulation, implementing the published selection, and estimating the predicted signal yield in the published regions. \ColliderBench{} focuses on validating the recast pipeline against a signal model already considered in the original publication. This is the hardest step of the workflow, since it requires the public toolchain to reproduce a known target despite missing experiment-internal information. Once validated, swapping in a new signal model is comparatively trivial. The full recasting workflow continues by repeating this validated pipeline across a grid of signal-model parameter points to set limits on new physics. This final step multiplies compute without testing new agentic capability, and so falls outside the benchmark scope.

\ColliderBench{} formalizes this workflow as an agent benchmark. Each task is anchored to a single LHC search paper and specifies a target signal model, a benchmark parameter point, a target observable or signal region, and a fixed output template. Given the paper, task specification, public documentation, and a containerized high-energy-physics software environment, the agent must construct an executable simulation and analysis pipeline and return predicted signal yields from its own recast. The benchmark therefore evaluates a complete scientific workflow, translating an ambiguous public analysis description into code, simulation choices, event selections, and numerical predictions.

\paragraph{Task definition} Formally, a \ColliderBench{} task defines a problem instance $x = (\mathcal{P}, s, \mathcal{O}, \mathcal{B}, \mathcal{T})$, where $\mathcal{P}$ is the target search paper, $s$ is the specified signal benchmark, $\mathcal{O}$ is the target observable or signal region, $\mathcal{B}=\{b_1,\ldots,b_K\}$ is the prescribed set of histogram bins or counting regions, and $\mathcal{T}$ denotes the available tool environment. The agent must produce a non-negative yield vector
\begin{equation}
\hat{y} = (\hat{y}_1,\ldots,\hat{y}_K) \in \mathbb{R}_{\geq 0}^K,
\end{equation}
where \(\hat{y}_k\) is the predicted signal yield in bin or region \(b_k\) and $K$ the total number of bins. In addition to this numerical output, the agent is required to provide the executable artifacts that produced it, including analysis code, simulation cards or scripts, intermediate data products when relevant, and a short report documenting methodological choices.

The benchmark compares $\hat{y}$ to reference yields $y^\star$ obtained from curated reference recasts or public validation records, depending on the task. The reference values and evaluation code are not exposed to the agent during the run. This design separates the public scientific problem specification from the private scoring target, and prevents the task from reducing to information retrieval from plots, tables, or other public records. The central difficulty is that the mapping $x \mapsto \hat{y}$ is not specified directly. The agent must reconstruct it from the paper and implement it through domain-specific, open-source scientific software.

\subsection{Task Corpus Design}
\label{sec:tasks}

\ColliderBench{} is built from public LHC searches for physics beyond the Standard Model. Each paper is treated as a source analysis from which several bounded task instances are derived. Each task fixes one target signal \(s\), one observable \(\mathcal{O}\) in a specified signal region, and one output template.

We select source analyses using two criteria.
\begin{itemize}
\item {\it Task tractability:}
The search must specify the relevant signal regions, object definitions, and event selections well enough for an external researcher with domain expertise to attempt the recast. The target signal must also be reproducible with reasonable fidelity using the public simulation tools available in the benchmark environment. This excludes analyses whose published results depend too strongly on internal detector reconstruction, specialized collaboration software, unavailable signal-generation machinery, or computationally prohibitive workflows.
\item{\it Reference availability:}
The search must provide quantitative reference results suitable for the evaluation. These may be published signal yields, binned signal distributions or plots that can be digitized reliably. We require a fixed reference target so that agent outputs can be scored automatically.
\end{itemize}

These criteria give the agent the public materials available to a human recaster.

The present release focuses on Supersymmetry (SUSY) simplified-model searches, which provide standardized signal topologies and public benchmark points while still requiring nontrivial reconstruction of the analysis logic (see Appendix~\ref{app:prose-to-code} for a simple example of such a reconstruction). Since a full LHC search often contains several signal models, mass points, event categories, and signal regions, \ColliderBench{} factorizes each source analysis into smaller tasks. Tasks derived from the same paper share common analysis ingredients but vary the target signal, mass point, observable, or region, helping separate paper-level understanding from signal-specific implementation and normalization errors.

The current task corpus uses CMS experimental searches at center-of-mass energy \(\sqrt{s}=13~\mathrm{TeV}\) with integrated luminosity \(\mathcal{L}_{\rm int}=35.9~\mathrm{fb}^{-1}\). The benchmark consists of 10 primary \texttt{Simulation} tasks, summarized in Table~\ref{tab:task-corpus}. Each task requires the agent to reproduce the binned signal yield for a fixed paper, signal benchmark, observable, and signal region. These 10 tasks define the headline benchmark and are used for the main model comparisons.

The task prompts are intentionally standardized across the corpus. Apart from paper-specific fields each prompt follows the same structure and gives the same type of instruction. This reduces prompt-format variability and makes differences in performance more attributable to the scientific content of the task rather than to changes in the agent interface. A representative prompt for one of the tasks is shown in Appendix~\ref{app:task-interface}.

\begin{table}[t]
    \centering
    \caption{Source analyses and simulation tasks in \ColliderBench{}.
    Each task fixes a paper, signal benchmark, target observable or
    signal region, and output template. Difficulty is graded from the our physicist-in-the-loop experiments: {\scriptsize$\bigstar$} (easy),
    {\scriptsize$\bigstar\bigstar$} (medium), {\scriptsize$\bigstar\bigstar\bigstar$} (hard).}
    \vspace{5pt}
    \scriptsize
    \label{tab:task-corpus}
    \setlength{\tabcolsep}{4pt}
    \resizebox{\textwidth}{!}{%
    \begin{tabular}{lllllll}
        \toprule
        Task & Analysis target & Signal \(s\) & obs. \(\mathcal{O}\) & LHC search & Reference & Diff. \\
        \midrule
        \rowcolor{gray!8}
        \texttt{sus-16-034\_sim-TChiWZ}      & leptons \(+\) jets & \texttt{TChiWZ}            & \(E_T^{\mathrm{miss}}\) & CMS-SUS-16-034 & \cite{CMS:2017kxn} & $\bigstar$ \\
        \texttt{sus-16-046\_sim-T5Wg}        & photons            & \texttt{T5Wg}              & \(S_T^{\gamma}\)        & CMS-SUS-16-046 & \cite{CMS:2017brl} & $\bigstar$ \\
        \texttt{sus-16-046\_sim-TChiWg}      & photons            & \texttt{TChiWg}            & \(S_T^{\gamma}\)        & CMS-SUS-16-046 & \cite{CMS:2017brl} & $\bigstar$ \\
        \rowcolor{gray!8}
        \texttt{sus-16-047\_sim-T5Wg\_highHT} & photons           & \texttt{T5Wg}, high-\(H_T\) & \(p_T^{\mathrm{miss}}\) & CMS-SUS-16-047 & \cite{CMS:2017qca} & $\bigstar\bigstar$ \\
        \rowcolor{gray!8}
        \texttt{sus-16-047\_sim-T5Wg\_lowHT}  & photons           & \texttt{T5Wg}, low-\(H_T\)  & \(p_T^{\mathrm{miss}}\) & CMS-SUS-16-047 & \cite{CMS:2017qca} & $\bigstar\bigstar\bigstar$ \\
        \rowcolor{gray!8}
        \texttt{sus-16-047\_sim-T6gg\_highHT} & photons           & \texttt{T6gg}, high-\(H_T\) & \(p_T^{\mathrm{miss}}\) & CMS-SUS-16-047 & \cite{CMS:2017qca} & $\bigstar\bigstar$ \\
        \rowcolor{gray!8}
        \texttt{sus-16-047\_sim-T6gg\_lowHT}  & photons           & \texttt{T6gg}, low-\(H_T\)  & \(p_T^{\mathrm{miss}}\) & CMS-SUS-16-047 & \cite{CMS:2017qca} & $\bigstar$ \\
        \texttt{sus-16-051\_sim-T2tt}         & single lepton     & \texttt{T2tt}              & \(E_T^{\mathrm{miss}}\) & CMS-SUS-16-051 & \cite{CMS:2017gbz} & $\bigstar$ \\
        \texttt{sus-16-051\_sim-T2tt\_comp}   & single lepton     & \texttt{T2tt}, compressed  & \(E_T^{\mathrm{miss}}\) & CMS-SUS-16-051 & \cite{CMS:2017gbz} & $\bigstar\bigstar\bigstar$ \\
        \texttt{sus-16-051\_sim-T2bW}         & single lepton     & \texttt{T2bW}              & \(E_T^{\mathrm{miss}}\) & CMS-SUS-16-051 & \cite{CMS:2017gbz} & $\bigstar\bigstar$ \\
        \bottomrule
    \end{tabular}}%
\end{table}

\subsection{Model Evaluation}
\label{sec:metrics}

\paragraph{Fidelity Metrics}
We first check whether the submission is valid. A valid submission contains the required output template, is syntactically well formed, and has finite non-negative values for all required bins. Runs that fail this check receive no numerical credit.

Our primary metric is the relative \(L^2\) distance between two binned histograms,
\begin{equation}
    d(\hat y,y^\star)
    =
    \sqrt{
    \frac{
    \sum_{k=1}^K
    \left(\hat y_k-y^\star_k\right)^2
    }{
    \sum_{k=1}^K
    y_k^{\star\,2}
    }}
    .
\end{equation}
This dimensionless metric measures the bin-level discrepancy relative to the reference histogram. By construction, discrepancies in high-yield bins contribute more strongly than discrepancies in sparsely populated bins. This is appropriate for recast validation, where reproducing the bulk of the selected signal distribution is typically more important than matching small tail fluctuations or low-yield bins that are more sensitive to modeling details. We also report a yield-normalization error. Defining \(\hat{Y}=\sum_k\hat{y}_k\) and \(Y^\star=\sum_k y^\star_k\), we compute
\begin{equation}
    \delta_{\rm norm}
    =
    \frac{|\hat{Y}-Y^\star|}{Y^\star}.
\end{equation}
This diagnostic helps separate failures in the relative distribution from failures in the absolute event-rate calculation.

For aggregate reporting, we convert the continuous error into a binary prediction accuracy,
\begin{equation}
    \mathrm{Acc}_{\tau}
    =
    \mathbb{I}\!\left[
    d_{\rm task}<\tau
    \right],
\end{equation}
where \(\tau\) is a fixed tolerance and $\mathbb{I}(\cdot)$ is the indicator function, equal to \(1\) when the condition inside the brackets is true and \(0\) otherwise.

\paragraph{Provenance Judge}
Numerical scoring is complemented by a provenance audit. Since some reference values may be available in public papers, plots, or auxiliary records, we use an LLM judge to inspect the agent workspace and execution trace. The judge does not assign the primary numerical score. Instead, it flags whether the submitted values are traceable to a legitimate simulation-and-analysis workflow, correspond to an incomplete run, or arise from an invalid shortcut. Specifically, the LLM judge returns three possible flags with the following definitions/specifications:

\begin{itemize}
    \item \textsc{Passed}. The submitted values are traceable to a genuine simulation-and-analysis attempt, even when the final numerical prediction is poor.

    \item \textsc{Failed}. No valid results were submitted. The output \texttt{yaml} file is left untouched, contains unresolved \texttt{null} values, or is otherwise not scorable. These cases typically arise when the agent abandons the workflow or reaches the time limit before producing a completed prediction.

    \item \textsc{Fabricated}. Non-null values are submitted, but they cannot be traced to a complete, agent-verified pipeline. The agent presents values as if they came from a genuine computation when they do not. These cases include hand-coded fallback arrays, surrogate shape generators, made-up efficiencies, or values copied from public sources.
\end{itemize}

\paragraph{Cost and Efficiency}
We report API cost, token usage, wall-clock time as separate efficiency metrics. These quantities are not folded into the fidelity score, since scientific accuracy and inference cost represent distinct axes of agent performance.

\section{Experiments}
\label{sec:experiments}

In this section we present the main benchmark results. Additional results, including cost and efficiency metrics, and Tool-ablation  are reported in Appendix~\ref{app:additional-results}.

\subsection{Experimental Setup}
\label{sec:experimental-setup}

We evaluate \ColliderBench{} using single-session autonomous coding agents. Each run starts from a fresh workspace containing the task prompt, target paper, output template, tool documentation, and containerized physics runtime. The agent may write code, execute commands, run the simulation stack, and revise its analysis before submitting a final output \texttt{yaml} template. Hidden reference event yields and evaluator code are not available to the agent during the run.


\paragraph{Agents}
We evaluate a capability ladder of agents through standard terminal interfaces. Anthropic models are run with Claude Code, OpenAI models with the Codex CLI, and DeepSeek-V4 with ForgeCode~\citep{forgecode2026} set to their maximal effort. Each model--harness pairing is treated as the evaluated system, since long-horizon scientific performance depends on planning, file editing, command execution, and recovery from errors in addition to the base model. All systems use the same sandbox, task materials, and scoring pipeline. All agents are given a wall-time budget of 2.5 hours per task and access to 128 AMD EPYC 7763 CPU cores for simulation workloads. 

\paragraph{Physicist-in-the-loop Baseline}
As a reference point, we include a supervised reproduction baseline using the latest Claude Code model, Opus 4.7, as a coding and analysis collaborator under the supervision of a human domain expert. This baseline is intended to calibrate the difficulty of the recasting workflow when a physicist remains responsible for scientific judgment.

Additional experimental details are given in Appendix~\ref{app:experimental-details}.

\begin{figure}[!t]
    \centering
    \includegraphics[width=1\textwidth]{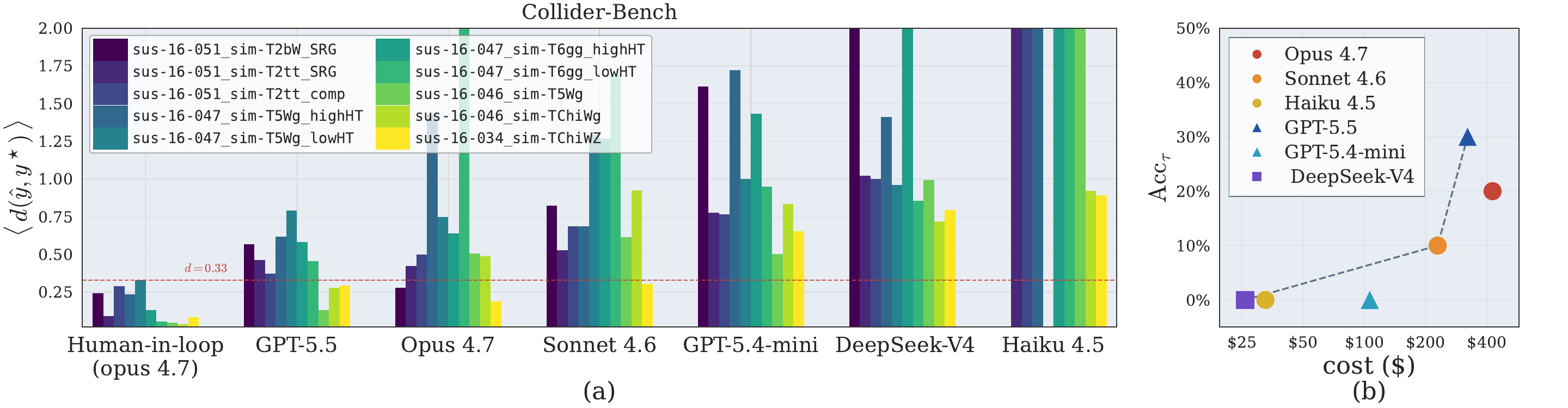}
\caption{
{\bf (a)} The mean relative \(L^2\) distance for each model and task (over 3 independent runs), with lower values indicating better agreement with the hidden reference yields. {\bf (b)}   The Pareto frontier of agent performance for \(\mathrm{Acc}_{\tau}\) versus inference cost for a fidelity threshold of $\tau=0.33$.
}\label{fig:pareto}
\end{figure}

\subsection{Main Results}
\label{sec:main-results}
The main evaluation covers the 10 primary \texttt{Simulation} tasks. We report the relative \(L^2\) fidelity metric and the thresholded acceptance rate \(\mathrm{Acc}_{\tau}\).
We set the validation threshold to \(\tau=0.33\), chosen as the worst relative \(L^2\) error achieved by the {\it physicist-in-the-loop} baseline across the task set. This gives a deliberately lenient success criterion: an autonomous agent is counted as passing only if it matches or improves on the weakest supervised reproduction.

Figure~\ref{fig:pareto} summarizes agent performance on the \texttt{Simulation} tasks of \ColliderBench{}. The left panel reports the mean relative \(L^2\) error for each model and task over three independent runs. The right panel shows \(\mathrm{Acc}_{\tau}\) versus average inference cost for \(\tau=0.33\).

The benchmark exposes a clear gap between autonomous agents and supervised recasting workflows. Autonomous agents improve along the model capability ladder and form a visible cost--performance frontier, but even the strongest systems pass only a subset of the tasks. Thus, current agents can sometimes execute substantial parts of the recast, but they do not yet match the reliability and judgment of an expert-supervised workflow.

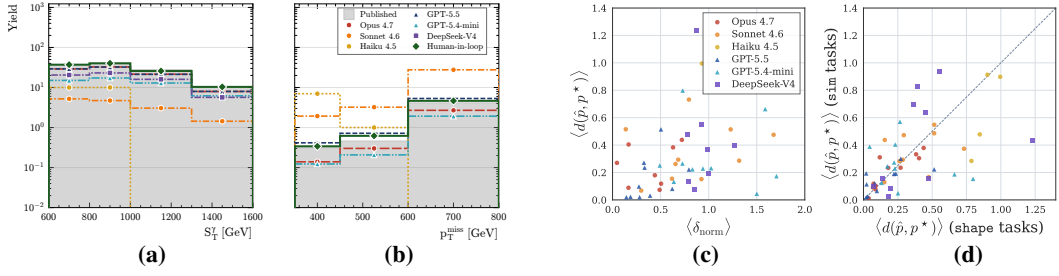
\begin{figure}[t]
    \centering

    \begin{tikzpicture}
        \node[inner sep=0] (leftimg)
        {\includegraphics[width=0.475\textwidth]{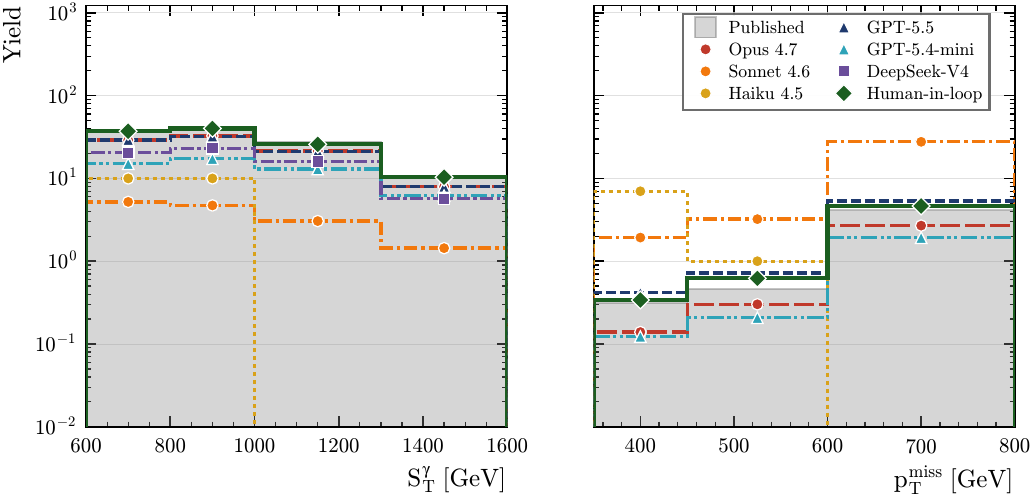}};

        \node[
            anchor=north west,
            font=\bfseries\small,
            fill=white,
            inner sep=1pt
        ] at ([xshift=17mm,yshift=-32mm]leftimg.north west) {(a)};

        \node[
            anchor=north west,
            font=\bfseries\small,
            fill=white,
            inner sep=1pt
        ] at ($(leftimg.north west)!0.52!(leftimg.north east)+(15mm,-32mm)$) {(b)};
    \end{tikzpicture}
    \hfill
    \begin{tikzpicture}
        \node[inner sep=0] (rightimg)
        {\includegraphics[width=0.475\textwidth]{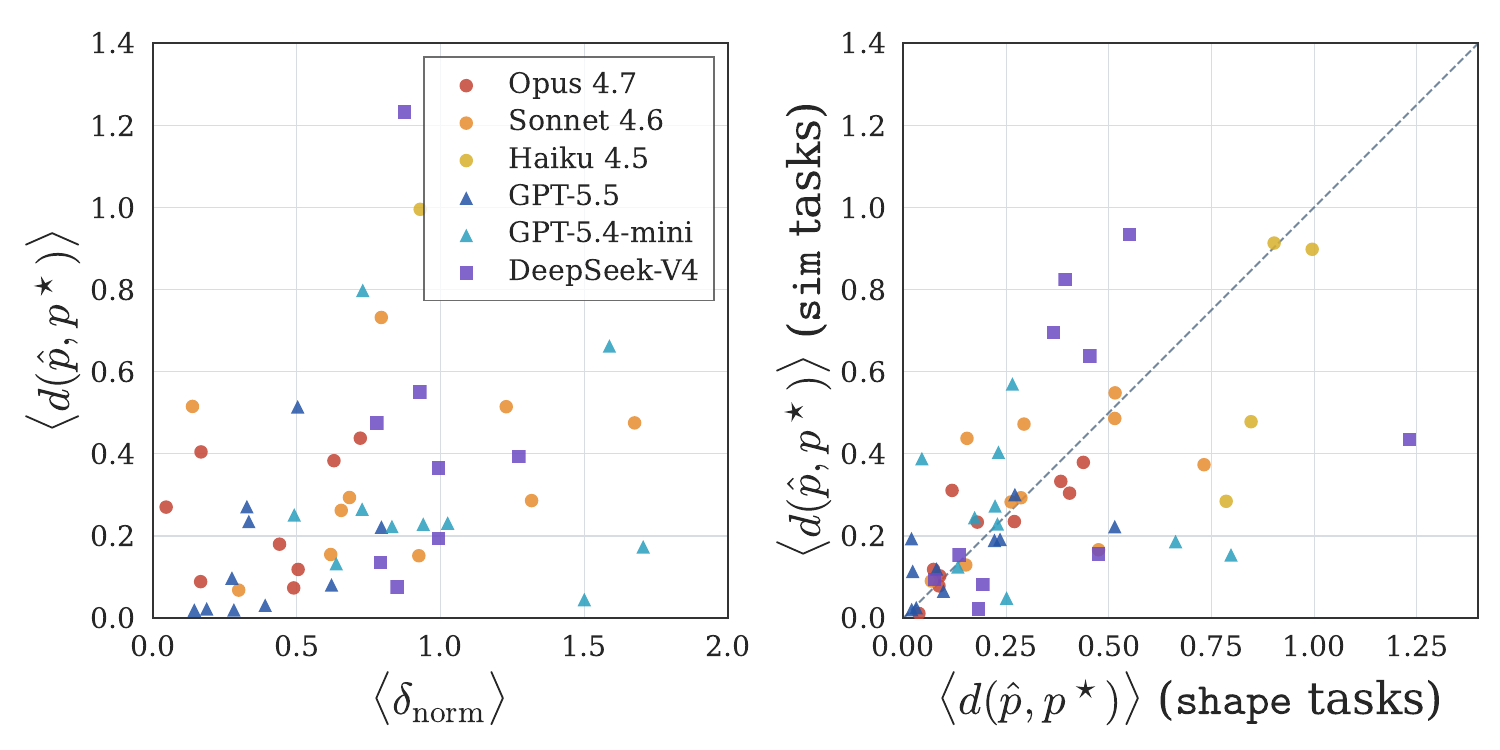}};

        \node[
            anchor=north west,
            font=\bfseries\small,
            fill=white,
            inner sep=1pt
        ] at ([xshift=17mm,yshift=-33mm]rightimg.north west) {(c)};

        \node[
            anchor=north west,
            font=\bfseries\small,
            fill=white,
            inner sep=1pt
        ] at ($(rightimg.north west)!0.52!(rightimg.north east)+(17mm,-33mm)$) {(d)};
    \end{tikzpicture}

    \caption{
Representative simulation-task predictions, for {\bf (a)}
\texttt{sus-16-034\_sim-TChiWZ} and and {\bf (b)} \texttt{sus-16-047\_sim-T6gg\_highHT}, overlaid against the published yields in gray. The dark green diamond-marked line is our physicist-in-the-loop reproduction, shown as a non-autonomous reference. For each agent we show the best of three runs according to the relative \(L^2\) distance \(d(\cdot,\cdot)\) on the absolute binned yields. The physicist-in-the-loop baseline is shown in green. The examples illustrate both near-reproductions with residual normalization offsets and catastrophic failures in which the predicted yield is concentrated in the wrong bins or falls below the displayed range. {\bf (c)} Correlations of relative $L^2$ distance with yield error. {\bf (d)}   Correlation of shape-only $L^2$ metrics between \texttt{Simulation} and \texttt{Shape} tasks.
}
    \label{fig:paper_sim_overlays}
\end{figure}

Performance is also strongly task dependent, exposing the intended difficulty hierarchy in the benchmark design. Some systems obtain low error on particular searches while failing on related tasks, indicating that success is not determined only by generic coding ability or toolchain access. Figure~\ref{fig:paper_sim_overlays} illustrates this heterogeneity with two representative tasks (see figure \ref{fig:appendix_sim} for the remaining tasks). On \texttt{sus-16-034\_sim-TChiWZ} (left panel), several stronger agents reproduce the qualitative shape of the published distribution, with discrepancies dominated by an approximately uniform normalization offset. On \texttt{sus-16-047\_sim-T6gg\_highHT}, the ordering changes and several models fail much more severely, including cases where the predicted yield is orders of magnitude too small or concentrated in the wrong bins. Thus, the corpus probes a spectrum of recasting difficulty rather than a collection of interchangeable histogram-filling problems.

\subsection{Shape--Normalization Diagnosis}
\label{sec:shape-normalization}

The simulation score alone does not reveal which component of the prediction failed. A poor simulation result may arise from an incorrect implementation of the event selection, an incorrect absolute normalization, or both. To separate these effects, we decompose the predicted event-yield vector as
\begin{equation}
    \hat{y}_k = \hat{Y}\,\hat{p}_k,
    \qquad
    \sum_{k=1}^K \hat{p}_k = 1,
\end{equation}
where \(\hat{p}_k\) is the normalized distribution over bins and \(\hat{Y}\) is the total predicted yield. Appendix~\ref{app:yield-normalization} explains how these quantities correspond to the ingredients of a collider recast.

We use \(d(\hat p,p^\star)\) to measure reconstruction of the normalized binned distribution independently of its absolute scale. As shown in Fig.~\ref{fig:paper_sim_overlays}(c), the shape error for each agent is not tightly correlated with the normalization error \(\delta_{\rm norm}\), or equivalently \(d(\hat y,y^\star)\), indicating that the two metrics can potentially capture distinct failure modes. Appendix~\ref{app:additional-results}, Table~\ref{tab:sim_shape_scores}, reports \(d(\hat p,p^\star)\) for all tasks and agents, while Table~\ref{tab:sim_scores} reports the corresponding full-yield errors. Overall, agents perform substantially better on shape reconstruction than on full yield reconstruction. This suggests that many failures arise not from the qualitative event-selection structure alone, but from converting a selected simulated sample into correctly normalized event yields.

The normalization component is a recurring bottleneck. It requires the agent to identify the appropriate production rate, combine it with the dataset size, account for particle decay assumptions, and propagate these choices consistently into the event weights. In several runs, agents handled the cross-section tool incorrectly, for example by using an inconsistent process or mass point, mixing normalization conventions, or failing to propagate the computed rate into the final histogram. Such errors can preserve an approximately correct histogram shape while shifting the total yield by a large factor. This exposes a failure mode that is largely invisible in code-only benchmarks: an agent may produce plausible analysis code and a qualitatively reasonable distribution while still failing the quantitative normalization required for a scientific prediction.

\subsection{Task Ablation}
\label{sec:task-ablation}

The full \texttt{Simulation} task is a compound scientific workflow. To produce a final binned yield, the agent must chain together several steps, including signal generation, showering, detector simulation, event selection, histogramming, and absolute normalization. A failure in the final prediction therefore does not by itself identify which subcomponent of the workflow is difficult for the agent. The shape--normalization decomposition in Sec.~\ref{sec:shape-normalization} provides a natural way to isolate one such component.

Motivated by this, we construct a simpler set of \texttt{Shape} tasks. These tasks use the same papers, signal benchmarks, observables, signal regions, and output binning as the primary \texttt{Simulation} tasks, but remove the requirement to predict the absolute event yield. The agent is instead asked to reproduce only the normalized shape of the binned distribution. This reduces the dependence on rate-normalization tools while still requiring the agent to identify the signal model and implement the event selection.

We run the same agents systems on this secondary task suite and evaluate them with the normalized shape metric \(d(\hat p,p^\star)\). The normalized shape metrics are compared between corresponding \texttt{Simulation}  and \texttt{Shape} tasks in Figure~\ref{fig:paper_sim_overlays} (d). We see that the two errors, especially for the best-performing agents, are strongly correlated, indicating that removing the normalization objective does not substantially change the underlying shape-reconstruction behavior. This supports the interpretation that shape extraction and absolute normalization are largely separable failure modes in the current benchmark. It also suggests a plausible path for future agent designs: a specialized normalization or cross-section-checking module could improve full simulation performance without requiring a wholesale change to the event-selection pipeline. For complete results from the \texttt{Shape} task, see Appendix~\ref{app:additional-results}.

\subsection{Provenance Test}
\label{sec:provenance_test}

Because numerical fidelity alone does not verify that a submitted histogram was produced by a valid recasting workflow, we additionally judge provenance using Claude-Opus-4.6 as an LLM judge with access to each agent's transcript and intermediate artifacts. Across 364 judged runs, we find the LLM judge returns \textsc{Passed} (\(87\%\)), \textsc{Failed} (\(6\%\)), \textsc{Fabricated} (\(6\%\)), more details can be found in Appendix~\ref{app:additional-results}. 
The provenance labels reveal model-dependent failure behavior:
\begin{itemize}
    \item Frontier models like Opus 4.7 and GPT-5.5 have near perfect pass rates producing scorable submissions in all judged runs. By contrast, failed runs from the remaining agents are usually traceable to incomplete pipelines, timeouts, malformed outputs, or incorrect but identifiable analysis choices.

    \item Fabrication is concentrated in smaller or lower-cost models. Haiku 4.5 accounts for the majority of fabricated submissions, with smaller contributions from DeepSeek-V4 and GPT-5.4-mini. Typical fabrication patterns include hard-coded fallback arrays, ad hoc probability-density samplers used in place of the Monte Carlo simulation chain, and references to published yields after runs that had selected zero events.

    \item Some passed runs still reveal time-budget limitations. For example, in one Opus 4.7 run, the agent correctly diagnosed during detector simulation that an invisible particle was being reconstructed as visible and patched the relabeling. However, the corrected pipeline did not finish before submission, so the final yields were taken from the broken first pass.

\end{itemize}

Overall, these cases show why \ColliderBench{} requires provenance-aware evaluation. A yield vector can receive a numerical score regardless of whether it was produced by a valid scientific workflow, an abandoned pipeline, or an invalid shortcut. The execution trace is therefore essential for distinguishing wrong but legitimate recasts from fabricated or incomplete submissions.

\section{Conclusion}
\label{sec:conclusion}
In this work, we introduced \ColliderBench{}, a benchmark designed to evaluate agentic coding systems on the task of reproducing collider physics analyses. A defining challenge of this benchmark is that publicly available information is insufficient to uniquely determine the correct solution, forcing models to make reasonable choices to fill the gaps. We evaluated six off-the-shelf agents on ten tasks spanning four LHC analyses, using quantitative metrics. Overall, we found that most agents successfully executed the assigned tasks, but no agent reliably matched the performance of a physicist-in-the-loop. We discussed several recurring pitfalls and observed a notable improvement in performance when tasks were restricted to well-scoped subsets of the original analysis. With \ColliderBench{}, we contribute a realistic and challenging testbed for probing state-of-the-art agentic workflows, and we plan to expand it by incorporating additional analyses over time.

\section*{Acknowledgments}
We are grateful to Sabine Kraml for discussions and early collaboration on this project. We thank Gregor Kasieczka and Aman Upadhyay for helpful discussions.
DAF, SPS, IP and DS were supported by DOE grant DOE-SC0010008. This research used resources of the National Energy Research Scientific Computing Center, a DOE Office of Science User Facility supported by the Office of Science of the U.S. Department of Energy under Contract
No. DE-AC02-05CH11231 using NERSC award HEPERCAP0027491.

\bibliographystyle{plainnat}
\bibliography{reference}

\clearpage
\appendix

\section{Brief Primer on Collider Physics}
\label{app:collider-primer}
\begin{figure}[!htbp]
    \centering
\includegraphics[width=1\textwidth]{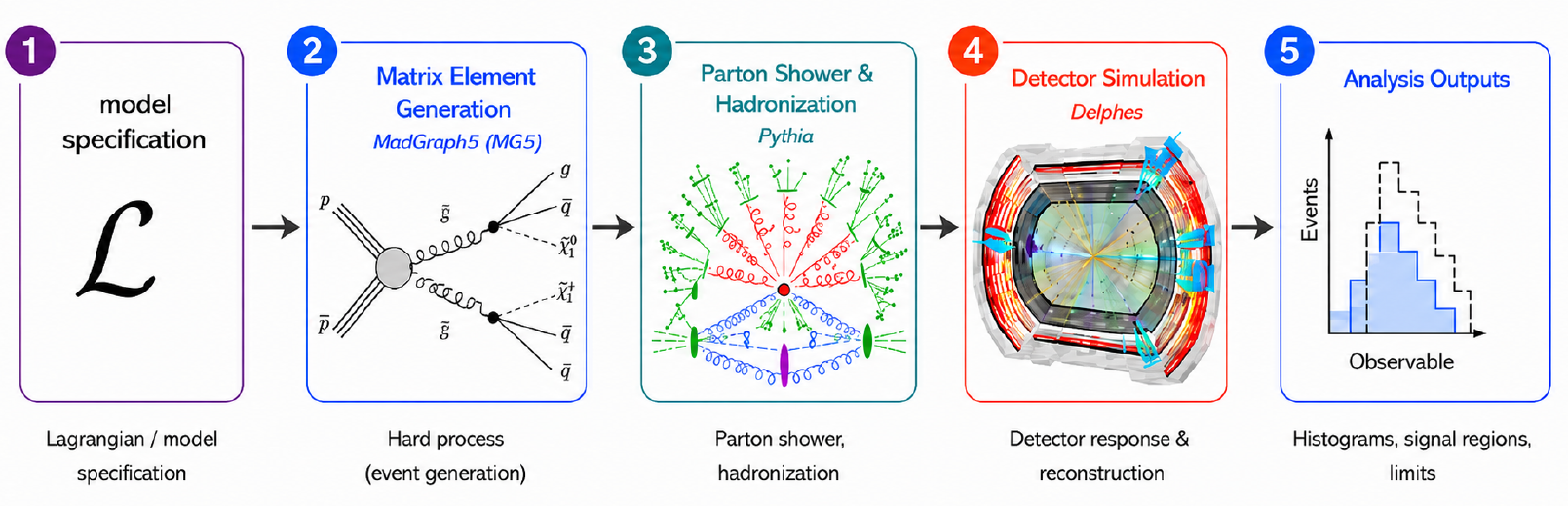}
\caption{Schematic of the event simulation pipeline. A Lagrangian $\mathcal{L}$ specifies the theoretical model (e.g. SUSY), successive stages simulate the hard scattering, parton showering and hadronization, and detector response, ultimately producing the observables compared to data.}
\label{fig:recast_pipeline}
\end{figure}
Collider physics studies the outcomes of high-energy particle collisions. At the Large Hadron Collider (LHC) at CERN, two beams of protons are accelerated in opposite directions and brought into collision at several points around the ring. Large general-purpose detectors such as ATLAS and CMS surround these collision points and record the particles produced in each collision. The raw detector signals are processed into reconstructed objects, such as electrons, muons, photons, jets, and missing transverse momentum. These reconstructed objects form the basic data record used in most LHC analyses.

A single proton--proton collision record is called an event. Each event contains a set of reconstructed objects and their measured kinematic properties, such as energy, momentum, and direction. An LHC search defines a set of rules for selecting events that may be compatible with a particular new-physics signal. These rules are usually based on kinematic quantities. For example, a search may require at least one high-energy photon, several energetic jets, and a large amount of missing transverse momentum. Events satisfying these requirements are grouped into signal regions, which are designed to have enhanced sensitivity to a possible signal.

The statistical structure of many searches is close to an event-counting experiment. The analysis predicts how many background events are expected in each signal region under the Standard Model, then compares this expectation to the number of events observed in data. A hypothetical new-physics model predicts an additional number of signal events. If the observed event counts are compatible with the background-only prediction, the analysis can place limits on the signal model. If there is a significant excess, the analysis may provide evidence for new physics.

The theoretical description of particle physics is specified analytically by a Lagrangian, which defines the particle content and interactions. To predict what a given theory would produce in the detector, physicists rely on a multi-stage Monte Carlo simulation pipeline, decomposed by energy scale, as schematically visualized in Figure~\ref{fig:recast_pipeline}. The first stage generates the hard scattering process at short distances, such as the production of a pair of new heavy particles. A second stage simulates parton showering and hadronization, which describe how the energetic quarks and gluons emerging from the hard process radiate further partons and ultimately bind into hadrons. These are reconstructed experimentally as jets. A final detector-simulation stage estimates how the generated particles would appear in a real detector. Full detector simulation is computationally expensive and is usually available only within the experimental collaborations, so external researchers rely on fast detector-simulation tools, which provide approximate detector responses at much lower cost.

A recast uses a published search to study a signal model beyond those explicitly analyzed in the original paper. Since the full internal analysis code is usually unavailable, the researcher must reconstruct the analysis from public information. This includes identifying the correct signal region, implementing the object definitions and event selections, generating simulated signal events, applying detector simulation, and estimating the final signal yield. \ColliderBench{} turns this workflow into a benchmark for AI agents.

\subsection{Recasting toolbox}
\label{app:recasting_toolbox}

Each task ships with a fixed set of command-line tools that wrap the public simulation stack and provide structured access to the public information sources a recaster needs. The agent uses these tools as it would in a real recasting workflow. Their interfaces are documented in the workspace. \texttt{read-paper} extracts text and figures from the target publication. \texttt{hepdata} queries HEPData~\citep{Maguire:2017ypu}, the standard repository for tabulated experimental results, for numerical material associated with the analysis. \texttt{cms-opendata} browses and streams datasets from the CMS Open Data portal. \texttt{feynrules} fetches UFO model files from the FeynRules database, which encode the analytical Lagrangian in the format consumed by the generator. \texttt{simulate} provides documentation and discovery utilities for the simulation stack. The underlying tools are invoked through their native command-line interfaces. MadGraph5~\citep{Alwall:2014hca}, Pythia 8.313~\citep{Bierlich:2022pfr} evolves the outgoing partons into parton showers and hadronises them into the stable particles that reach the detector; and Delphes 3.5.0~\citep{deFavereau:2013fsa} applies a fast, parametric model of the detector response, returning reconstructed objects (jets, leptons, photons, missing transverse momentum). \texttt{prospino} computes cross-section corrections with Prospino 2.1~\citep{Beenakker:1996ch}, used to normalise simulated yields. \texttt{run-analysis} executes the agent's analysis script inside the benchmark's Python environment.

\subsection{Glossary}

\begin{description}
    \item [Event:] A reconstructed record of a single proton--proton collision. It contains physics objects such as photons, leptons, jets, and missing transverse momentum.

    \item[Recasting:]
    The process of reusing a published search to test a signal model different from those probed in the original analysis.

    \item[Signal:]
    The event contribution predicted by a hypothesized new-physics model.

    \item[Signal region:]
    A subset of events selected to enhance sensitivity to a target signal. Signal regions are usually defined by cuts on reconstructed objects and kinematic variables.

    \item[\bf Physics object:] A reconstructed particle-like object used in an analysis. Common examples are electrons, muons, photons, jets, and tagged jets.

    \item[\bf Jet:] A collimated spray of hadrons produced when an energetic quark or gluon is emitted in a collision. Jets are among the most common objects in LHC analyses.

    \item[\bf Lepton:]
    A particle such as an electron or muon. In collider analyses, leptons are useful because they can often be reconstructed cleanly.

    \item[\bf Photon:]
    A particle (quanta) of light. Some new-physics searches target final states containing energetic photons.

    \item[Missing transverse momentum $p_T^{\rm miss}$:]
    A vector quantity that measures momentum imbalance in the plane transverse to the proton beams. It is often interpreted as evidence for invisible particles escaping the detector, such as neutrinos or hypothetical dark-matter candidates.

    \item[Background:]
    Standard Model processes that can mimic the signal in the detector. Searches compare observed data to the expected background and possible signal contributions.

    \item[Cross section $\sigma$:]
    A measure of how frequently a physical process occurs in proton--proton collisions. Larger cross sections correspond to more common processes.

    \item[Branching fraction $\beta$:]
    The probability that an unstable particle decays through a particular channel. If a signal can decay in several ways, the relevant branching fraction affects the expected event yield.

    \item[Integrated luminosity $\mathcal{L}_{\rm int}$:]
    A measure of the size of the recorded dataset. Larger integrated luminosity means more collisions were recorded. For a process with cross section \(\sigma\), the expected number of produced events is approximately \(\mathcal{L}_{\rm int}\sigma\), before accounting for decays, detector effects, and selection cuts.

    \item[Acceptance :]
    The fraction of generated signal events that fall within the kinematic and geometric region targeted by the analysis.

    \item[Efficiency:]
    The probability that an event passing the idealized particle-level requirements is successfully reconstructed and selected by the detector-level analysis.

    \item[Acceptance times efficiency:]
    The total fraction of generated signal events that survive the analysis selection. This is often the central quantity measured by a recast.

    \item[Monte Carlo simulation/generation:]
    A computational method for generating synthetic collision events from a physical model. In Collider-Bench, agents use Monte Carlo tools to produce signal samples.

    \item[Parton shower:]
    A simulation of the radiation emitted by energetic quarks and gluons after the short-distance collision.

    \item[Hadronization:]
    The process by which quarks and gluons turn into color-neutral hadrons. This cannot be computed exactly from first principles in most collider simulations and is modeled phenomenologically.

    \item[Detector simulation:]
    A simulation of how particles interact with the detector and how they are reconstructed as physics objects.

    \item[Fast detector simulation:]
    An approximate detector simulation that is much cheaper than the full simulation used by experimental collaborations. Delphes is a common example.

    \item[Recast validation:]
    The process of checking whether an external implementation of a published analysis can reproduce known signal yields or distributions from the original analysis.
\end{description}

\begin{figure}[!h]
    \centering
    \begin{minipage}[t]{0.50\linewidth}
        \begin{tcolorbox}[
            enhanced,
            equal height group=prosecode,
            colback=paperbg,
            colframe=framecol,
            boxrule=0.5pt,
            arc=2mm,
            left=1.2mm,right=1.2mm,top=1mm,bottom=1mm,
            boxsep=1mm
        ]
        \footnotesize
        Events are selected if they contain at least one photon with
        \(p_T > 100~\mathrm{GeV}\) in the EB with \(|\eta| < 1.4442\).
        To reliably predict the background, the photon is not allowed to be
        parallel or anti-parallel to \(\vec{p}_T^{\,\mathrm{miss}}\) within an
        azimuthal angle of
        \(|\Delta\phi(\pm \vec{p}_T^{\,\mathrm{miss}}, \vec{p}_T^{\,\gamma})| < 0.3\).
        \end{tcolorbox}
    \end{minipage}
    \hfill
    \begin{minipage}[t]{0.46\linewidth}
        \begin{tcolorbox}[
            enhanced,
            equal height group=prosecode,
            colback=white,
            colframe=framecol,
            boxrule=0.5pt,
            arc=2mm,
            left=0mm,right=0mm,top=0mm,bottom=0mm,
            boxsep=0.5mm
        ]
\begin{lstlisting}[
    style=githubpython,
    basicstyle=\ttfamily\tiny\color{ghname},
    escapeinside={(*@}{@*)}
]
(*@\color{blue}\textit{// pseudocode:}@*)
photons = [
    ph for ph in event.photons
    if ph.pt > 100 and abs(ph.eta) < 1.4442 and ph.is_in_EB
]
if not photons:
    reject(event)

lead_photon = max(photons, key=lambda ph: ph.pt)
dphi = abs(delta_phi(event.met_phi, lead_photon.phi))

if dphi < 0.3 or abs(math.pi - dphi) < 0.3:
    reject(event)
\end{lstlisting}
        \end{tcolorbox}
    \end{minipage}
    \caption{The agent must convert event-selection criteria from the published analysis description (left) into executable analysis logic (right).}
    \label{fig:prose-to-code}
\end{figure}

\section{Benchmark Construction Details}
\label{app:benchmark-construction}

\subsection{Event-selection specification to code}
\label{app:prose-to-code}

Figure~\ref{fig:prose-to-code} illustrates the translation step required in \ColliderBench{}. The agent must convert event-selection criteria from the published analysis description into executable analysis logic.

\subsection{Task Interface}
\label{app:task-interface}

Each \ColliderBench{} task is represented by the target paper \(\mathcal{P}\) in PDF format, a natural-language task prompt \texttt{TASK.md}, tool documentation \texttt{TOOLS.md}, and a null-filled output template in \texttt{yaml} format. The prompt specifies the target paper, signal benchmark, observable, and signal region. The output template specifies the bins or counting regions that the agent must fill at the end of the run. Hidden reference yields and evaluator code are not present in the agent workspace.

The sandbox controller pushes the following prompt to the freshly spawned agent:\\

{\tt \small
        You are a LHC experimentalist with expertise in Standard Model and BSM
        search strategies, event generation tools, event selection design, and
        statistical interpretation of collider data.
        Read these in order:
        \begin{itemize}
        \item 1. agent\_context/TASK.md   — the benchmark's task for this run.
        \item 2. agent\_context/AGENTS.md — your role and how to work.
        \item 3. agent\_context/TOOLS.md  — CLI tool reference.
        \end{itemize}
        Everything you need is in this directory.
}

\vspace{10pt}

\begin{agentcard}[\texttt{TASK.md}]
\small
\textbf{Paper:} CMS-SUS-16-046 \\
\textbf{Centre-of-mass energy:} 13 TeV \\
\textbf{Luminosity:} \(35.9~\mathrm{fb}^{-1}\) \\
\textbf{Task type:} simulation \\
\textbf{Signal benchmark:} \texttt{T5Wg\_1750\_1700} \\
\textbf{Observable:} \(S_T^\gamma\)
\medskip

\textbf{Task}

Implement the search analysis described in \textbf{CMS-SUS-16-046} and use it to predict the binned differential signal yield in $S_T^\gamma$ for the benchmark point \texttt{T5Wg\_1750\_1700}, in the analysis's {\bf signal region}, normalized to 35.9 $\mathrm{fb}^{-1}$ at $\sqrt s = 13$ TeV.\\

The agent should:
\begin{enumerate}
    \item Generate \texttt{T5Wg\_1750\_1700} events using a matrix-element generator, parton shower, and detector simulation chain of its choice.
    \item Read the paper to determine the object identification, event-selection requirements, and signal-region cuts that define the analysis, then apply them to the generated events.
    \item Histogram the surviving events in \(S_T^\gamma\) using the bin edges already present in the \texttt{results/*.yaml} template. The bin edges must not be modified.
\end{enumerate}

\textbf{Definitions}

\begin{itemize}
    \item \(S_T^\gamma\) -- the scalar sum of \(p_T^{\mathrm{miss}}\) and the transverse momenta of all photons in the event.
    \item \texttt{T5Wg\_1750\_1700} -- pair-produced gluinos at $m(\tilde g) = 1750$ GeV, each decaying via the {\tt T5Wg} simplified-model topology to a mass-degenerate wino NLSP at $m(\widetilde W) = 1700$ GeV` and a massless LSP.

\end{itemize}

\textbf{Output requirements}

\begin{center}
\begin{tabular}{ll}
\toprule
Artifact & Purpose \\
\midrule
\texttt{results/*.yaml} & Fill null bin values with predicted relative signal bin contents. \\
\texttt{analysis/*.py} & Event-selection code, runnable on the generated sample. \\
\texttt{data/*.root}, \texttt{sims/*.dat} & Selected-event files and generator/detector cards. \\
\texttt{report.md} & Methodological choices and deviations from the paper. \\
\bottomrule
\end{tabular}
\end{center}

\textbf{Important}

The goal is to predict the signal event distribution from your own simulation and analysis pipeline. Do not extract bin values from the paper's figures, tables, HEPData record, or elsewhere.

\end{agentcard}

\vspace{10pt}

\begin{agentcard}[\texttt{AGENTS.md}]
\small
\textbf{AGENTS}

You are analyzing CMS paper \texttt{\{arxiv\_id\}}.

\medskip
\textbf{What you have}
\begin{itemize}
    \item \texttt{papers/\{arxiv\_id\}.pdf} --- the paper. Read it with \texttt{bin/read-paper}.

    \item \texttt{results/*.yaml} --- histogram template with null values. The file contains two YAML documents separated by \texttt{---}: a metadata block with \texttt{instructions}, \texttt{description}, \texttt{target}, \texttt{cm\_energy\_gev}, and \texttt{luminosity\_fb}, followed by the HEPData-style histogram with \texttt{dependent\_variables} and \texttt{independent\_variables}. Fill the nulls in \texttt{dependent\_variables[0].values} in place. Do not modify bin edges or metadata.

    \item \texttt{object\_efficiencies/} --- detector efficiency files, if provided for the task.

    \item \texttt{bin/} --- command-line tools. A cheat sheet is shown below; the full reference is in \texttt{TOOLS.md}.

    \item \texttt{tools/} --- Python libraries and collider-physics toolkit.
\end{itemize}
\medskip
\textbf{Tools cheat sheet}

\begin{lstlisting}[basicstyle=\ttfamily\scriptsize,breaklines=true]
bin/read-paper papers/{arxiv_id}.pdf [--pages 3-5] [--figures]
bin/hepdata find <arxiv-id>
bin/hepdata get <inspire-id> "Table 1" --json
bin/cms-opendata search "ZZTo4L" --json
bin/cms-opendata files <recid> --json
bin/cms-opendata stream <root://url> --branches Muon_pt Muon_eta
bin/cms-opendata sample-info <recid>
bin/run-analysis
bin/feynrules list --search "vector-like quark"
bin/feynrules info <model>
bin/feynrules fetch <model> --extract --dest sim/models
bin/simulate info
bin/simulate --doc
mg5_aMC <proc_card>.dat
DelphesHepMC3 "$DELPHES_DIR/cards/delphes_card_CMS.tcl" out.root events.hepmc
# Pythia: write a small Python driver (`import pythia8`)
bin/prospino list-processes
bin/prospino help-process <proc>
bin/prospino run --process <proc> --sqrts 13000 --order <fixed-order> --slha <file>.slha
\end{lstlisting}
\end{agentcard}

\vspace{10pt}

\begin{agentcard}[\texttt{TOOLS.md}]
\small
\textbf{TOOLS}

All tools live in \texttt{bin/}. Call them with relative paths. Use the wrappers below, and fall back to the Python APIs listed at the bottom.

Each tool has a detailed reference that can be pulled with \texttt{bin/<tool> --doc}. This file is the index describing what each tool does and when to use it.

\medskip
\textbf{Tool index}
\begin{center}
\scriptsize
\begin{tabularx}{0.98\linewidth}{lXl}
\toprule
Tool & Purpose & Deep doc \\
\midrule
\texttt{read-paper}
& Extract paper text and render figures as PNGs.
& \texttt{bin/read-paper --doc} \\

\texttt{hepdata}
& Query the HEPData repository for published tables.
& \texttt{bin/hepdata --doc} \\

\texttt{cms-opendata}
& Browse and stream CMS Open Data, including data, MC, cross sections, and cards.
& \texttt{bin/cms-opendata --doc} \\

\texttt{simulate}
& Discover the MG5, Pythia, and Delphes stack and retrieve simulation guidance. The underlying CLIs are called directly.
& \texttt{bin/simulate --doc} \\

\texttt{feynrules}
& Browse and fetch UFO models from the FeynRules wiki.
& \texttt{bin/feynrules --doc} \\

\texttt{prospino}
& Compute NLO cross sections for SUSY pair production.
& \texttt{bin/prospino --doc} \\

\texttt{run-analysis}
& Execute \texttt{analysis.py} under the benchmark environment.
& \texttt{bin/run-analysis --doc} \\
\bottomrule
\end{tabularx}
\end{center}
\medskip
\textbf{When to pull \texttt{--doc}}

The one-line purposes above are usually enough to choose the right tool. Pull \texttt{bin/<tool> --doc} only when you are ready to invoke a tool and need:
\begin{itemize}
    \item a specific CLI flag or option not covered by the cheat sheet;
    \item the exact output schema, for example when parsing JSON;
    \item a tool-specific caveat or explanation for unexpected behavior.
\end{itemize}

The per-tool docs should be treated like man pages: useful when needed, but not required pre-reading.

\medskip
\textbf{Calling the tools}
\begin{itemize}
    \item Never run \texttt{python3 analysis.py} directly. Use \texttt{bin/run-analysis} so the correct conda environment is active.

    \item Never background \texttt{bin/run-analysis}, either with \texttt{run\_in\_background} or \texttt{\&}. It has a four-hour internal timeout, so blocking execution is safe.

    \item Always use \texttt{root://} XRootD URLs with \texttt{cms-opendata stream}, not \texttt{https://}.
\end{itemize}

\medskip
\textbf{Python modules}
\begin{itemize}
    \item HEP analysis stack: \texttt{uproot}, \texttt{awkward}, \texttt{numpy}, \texttt{hist}, \texttt{mplhep}, \texttt{yaml}.
\end{itemize}
\end{agentcard}

\subsection{Representative Output Template}
\label{app:output-template}

Each task provides a null-filled YAML output template. The agent must fill only the
\texttt{value: null} entries in the dependent variable while preserving the
metadata, bin edges, and file structure.

\begin{filecard}{taskbg}{taskframe}{tasktitle}{\texttt{results/histogram.yaml}}

\begin{lstlisting}[style=yamlplain]
instructions: "Fill the 'null' bin values in dependent_variables with your results.
               In the rightmost bin include ALL events with S_T^gamma > 1300 GeV.
               Do not modify bin edges.
              "
description: "Signal event counts for binned S_T^gamma"
target: "T5Wg, gluino 1750 GeV, NLSP 1700 GeV, Massless LSP"
cm_energy_gev: 13000
luminosity_fb: 35.9
---
dependent_variables:
- header: {name: T5Wg_1750_1700}
  values:
  - {value: null}
  - {value: null}
  - {value: null}
  - {value: null}
independent_variables:
- header: {name: STGAMMA, units: GEV}
  values:
  - {low: 600.0, high: 800.0}
  - {low: 800.0, high: 1000.0}
  - {low: 1000.0, high: 1300.0}
  - {low: 1300.0, high: 1600.0}
\end{lstlisting}

\end{filecard}

\subsection{From Shape to Absolute Event Yields}
\label{app:yield-normalization}

We decompose the agent's predicted yield in bin or region \(b_k\) as
\begin{equation}
    \hat{y}_k = \hat{Y}\,\hat{p}_k,
    \qquad
    \sum_{k=1}^K \hat{p}_k = 1.
\end{equation}
This decomposition separates two ingredients of a collider recast. The vector \(\hat{p}\) describes the relative distribution of selected signal events across the analysis bins. The scalar \(\hat{Y}\) fixes the absolute number of signal events expected in the dataset.

In collider language, the absolute event yield is determined by the size of the dataset, the rate at which the signal process is produced, the relevant decay probabilities, and the fraction of simulated events that survive the analysis selection. Schematically, one may write
\begin{equation}
    \hat{y}_k
    \simeq
    \mathcal{L}_{\rm int}\,
    {\sigma}_{\rm sig}\,
    \beta\,
    {(\mathcal{A}\epsilon)}_k ,
\end{equation}
where \(\mathcal{L}_{\rm int}\) is the integrated luminosity of the dataset (fixed for each search), \({\sigma}_{\rm sig}\) is the signal production cross section used by the agent (specific to each signal process), \(\beta\) denotes the relevant branching-fraction (specific to every particle decay channel), and \({(A\epsilon)}_k\) is the estimated acceptance times efficiency for bin or region \(b_k\).

The quantity \({(\mathcal{A}\epsilon)}_k\) is obtained from the agent's simulated events. If the agent generates \(N_{\rm gen}\) signal events and \(N_k\) of them pass the analysis selection and fall in bin \(b_k\), then in the simplest unweighted case
\begin{equation}
    {(\mathcal{A}\epsilon)}_k = \frac{N_k}{N_{\rm gen}}.
\end{equation}
For weighted Monte Carlo samples, this ratio is replaced by the corresponding weighted sum. In either case, \({(\mathcal{A}\epsilon)}_k\) measures how much of the generated signal survives the detector simulation, object reconstruction, event selection, and bin assignment.

The connection to the main-text decomposition follows by separating the total selected rate from its bin-by-bin distribution. Define
\begin{equation}
    \mathcal{A}\epsilon_{\rm tot}
    =
    \sum_{k=1}^K {(\mathcal{A}\epsilon)}_k .
\end{equation}
Then
\begin{equation}
    \hat{Y}
    =
    \mathcal{L}_{\rm int}\,
    {\sigma}_{\rm sig}\,
    \beta\,
    \mathcal{A}\epsilon_{\rm tot},
    \qquad
    \hat{p}_k
    =
    \frac{{(\mathcal{A}\epsilon)}_k}
         {\sum_{j=1}^K {(\mathcal{A}\epsilon)}_j}.
\end{equation}
Thus, shape tasks primarily evaluate the agent's estimate of the normalized acceptance pattern \(\hat{p}_k\), while simulation tasks also evaluate the absolute scale \(\hat{Y}\).

This distinction is useful because the two sources of error are physically different. An agent can obtain a reasonable shape while assigning the wrong total rate if it implements the event selection correctly but uses an incorrect luminosity, cross section, branching fraction, or event-weight normalization. Conversely, an agent can use the correct cross section and luminosity but still fail to predict the correct distribution if it misreads the object definitions, applies the wrong cuts, or assigns events to the wrong signal-region bins.

In practice, several additional details may enter the normalization. Published signal rates may use higher-order perturbative calculations, for example NLO or NLO+NLL cross sections, rather than the leading-order rate produced directly by an event generator. Analyses may also include filter efficiencies, generator-level branching assumptions, or model-specific decay tables. \ColliderBench{} simulation tasks test whether the agent can identify and apply these ingredients well enough to convert its simulated sample into the event yields expected for the published dataset.

\subsection{Agent Workspace and Sandboxing}
\label{sec:workspace-sandbox}

A \ColliderBench{} run is an interactive scientific workspace rather than a static question-answering instance. At the start of each task, the agent is placed in a fresh workspace containing the public problem specification and tools. The workspace also provides writable directories where the agent may inspect files, write and execute code, call the benchmark command-line tools, run the simulation stack, and revise its analysis before submitting a final prediction.

\paragraph{Controlled information surface and provenance.}
During scored runs, hidden reference yields, evaluator code, private scoring metadata, other tasks, and prior run artifacts are not mounted inside the agent workspace. The agent may use public documentation and auxiliary records to recover analysis metadata, such as bin definitions or validation information, but the submitted yields must come from its own recast rather than from copying target values from public tables, plots, or records.

Each run records its task identifier, configuration, workspace contents, generated artifacts, and final outputs. \ColliderBench{} also records an execution trace {\tt session.jsonl} of the agent's interaction with the environment, including commands, tool calls, intermediate outputs, and state changes observed during the run. This trace is not used to compute the primary numerical score, but it provides a provenance record for downstream qualitative evaluation. In particular, it enables an LLM judge or human reviewer to inspect whether the submitted prediction arose from a legitimate recasting workflow, a shortcut, or a failure mode not visible from the final yield vector alone.

\section{Experimental Details}
\label{app:experimental-details}

\subsection{Agent Systems}
\label{app:agent-systems}

We evaluate complete LLM models enhanced with agentic scaffolds rather than isolated language models. Anthropic models are run through Claude Code, OpenAI models through the Codex CLI, and DeepSeek-V4 through ForgeCode~\citep{forgecode2026}. ForgeCode~\citep{forgecode2026} is an open-source terminal-native coding harness that has recently performed strongly on terminal-based agent benchmarks.

All agents are launched through the same \ColliderBench{} sandbox controller that constructs a fresh workspace for each task, mounts the task inputs, starts the selected agent CLI inside the sandbox, records the execution trace, and collects the final output artifacts. We do not tune task prompts separately for each model. All agents receive the same public task specification and tool documentation.

\subsection{Run Protocol}
\label{app:run-protocol}

Each task run begins from an empty workspace containing the target paper, \texttt{TASK.md}, \texttt{TOOLS.md}, and the null-filled \texttt{yaml} output template. A run terminates when the agent voluntarily ends its turn, when it exhausts the allotted wall-clock budget (2.5 hours), or when the sandbox controller reports a fatal error; whatever output template exists at that moment is the submission, regardless of whether the agent intended it to be final.

A submission is considered \emph{passed} if the required output template exists, parses as YAML, has the expected bin structure, contains finite non-negative values and passes the LLM-judge provenance audit. Only submissions that pass these structural checks are scored with the relative-$L^2$ fidelity metric and the task-level validation indicator $\mathrm{Acc}_\tau$ defined in Sec.~\ref{sec:metrics}.

We do not delete failed runs from the cohort: submissions that are syntactically well-formed but whose bins are entirely \texttt{null}, all zero, or follow a non-standard schema the benchmark scorer cannot read are still entered into the per-model averages, so that the headline numbers reflect the run's actual outcome rather than silently removing pipeline failures. Runs flagged \textsc{fabricated} by the provenance audit are excluded from per-task aggregates, because their submitted values do not correspond to any agent-executed pipeline. Cost is computed from recorded model usage and reported per task.

\subsection{Physicist-in-the-loop Baseline}
\label{app:human-in-loop}

As a reference point, we include a supervised reproduction baseline using Claude Code as a coding and analysis collaborator under the supervision of a domain expert. Each supervised reproduction follows the same external-recast pipeline as the autonomous tasks. The agent generates parton-level signal events with {\tt MadGraph5\_aMC@NLO}, showers and hadronizes them with {\tt Pythia8}, applies detector simulation with {\tt Delphes}, processes the resulting event files, implements the published event selection, fills the analysis bins, and compares the final yields to the reference values.

The human role is supervisory rather than implementational. The agent writes most of the analysis code, including event loops, plotting scripts, cutflow tables, cross-section lookups, and yield-comparison utilities. The human provides the next analysis step, checks physical plausibility, and intervenes on domain-specific judgment calls such as object definitions, cross-section choices, event-weight normalization, and whether a discrepancy should be attributed to detector modeling or to an implementation error.

This baseline illustrates a central difficulty of collider-analysis reproduction. A capable agent can carry much of the coding burden when guided by an expert, but the bottleneck often lies in scientific judgment about conventions, normalizations, and which discrepancies matter.

\section{Additional Results}
\label{app:additional-results}

This appendix contains the following additional figures and tables.

\begin{figure}
    \centering
    \includegraphics[width=1\linewidth]{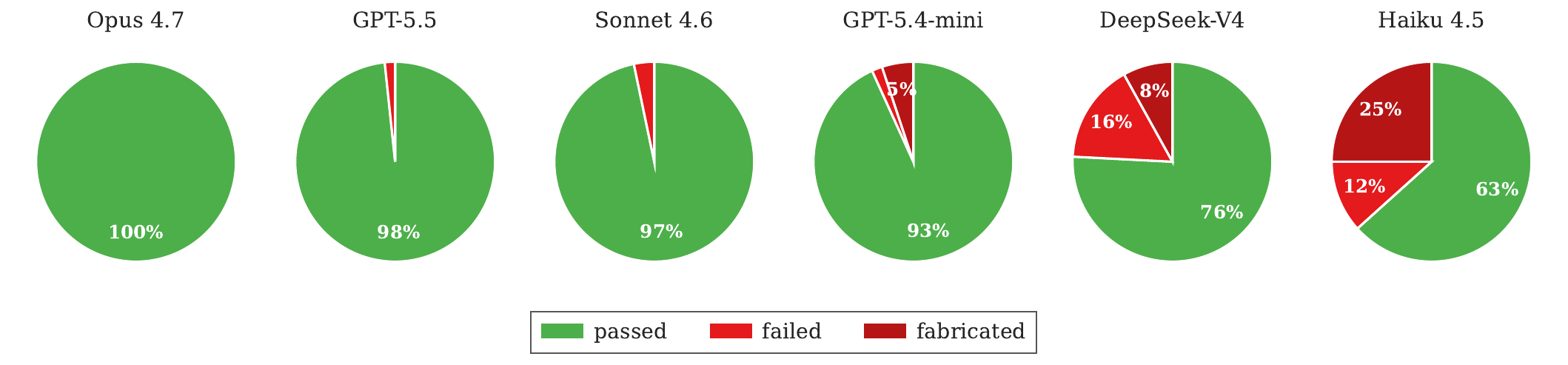}
    \caption{fraction   pass/fail runs.}
    \label{fig:appendix_pies}
\end{figure}

\begin{figure}[t]
    \centering
    \includegraphics[width=1\textwidth]{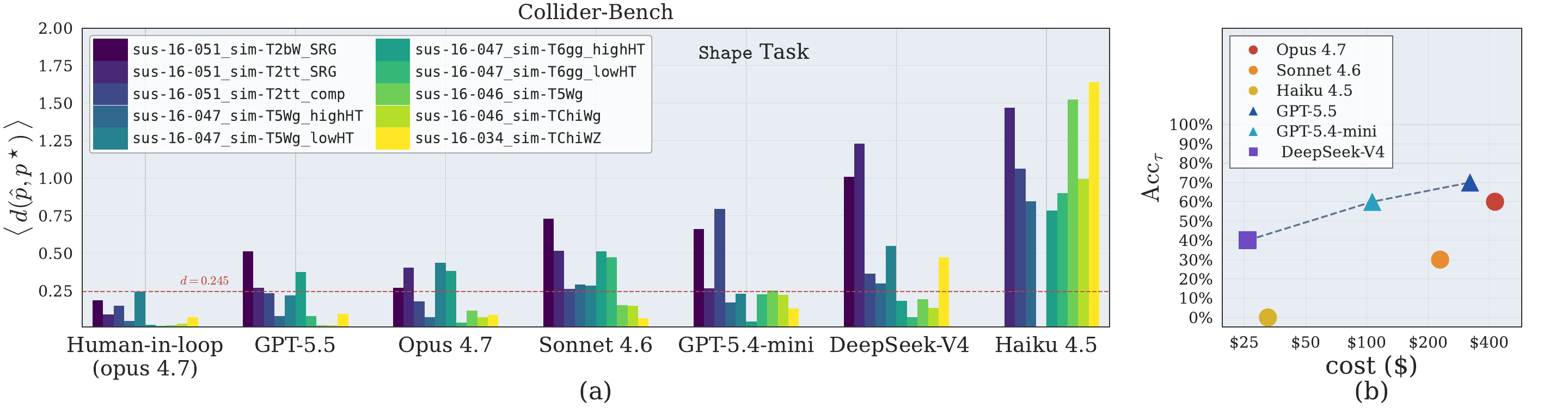}
\caption{
{\bf (a)} The mean relative \(L^2\) distance for each model and {\tt Shape} task (over 3 independent runs), with lower values indicating better agreement with the hidden reference yields. {\bf (b)} The Pareto frontier of agent performance for \(\mathrm{Acc}_{\tau}\) versus inference cost for a fidelity threshold of $\tau=0.33$.
}\label{fig:pareto_shape}
\end{figure}

\begin{figure}
    \centering
    \includegraphics[width=1\linewidth]{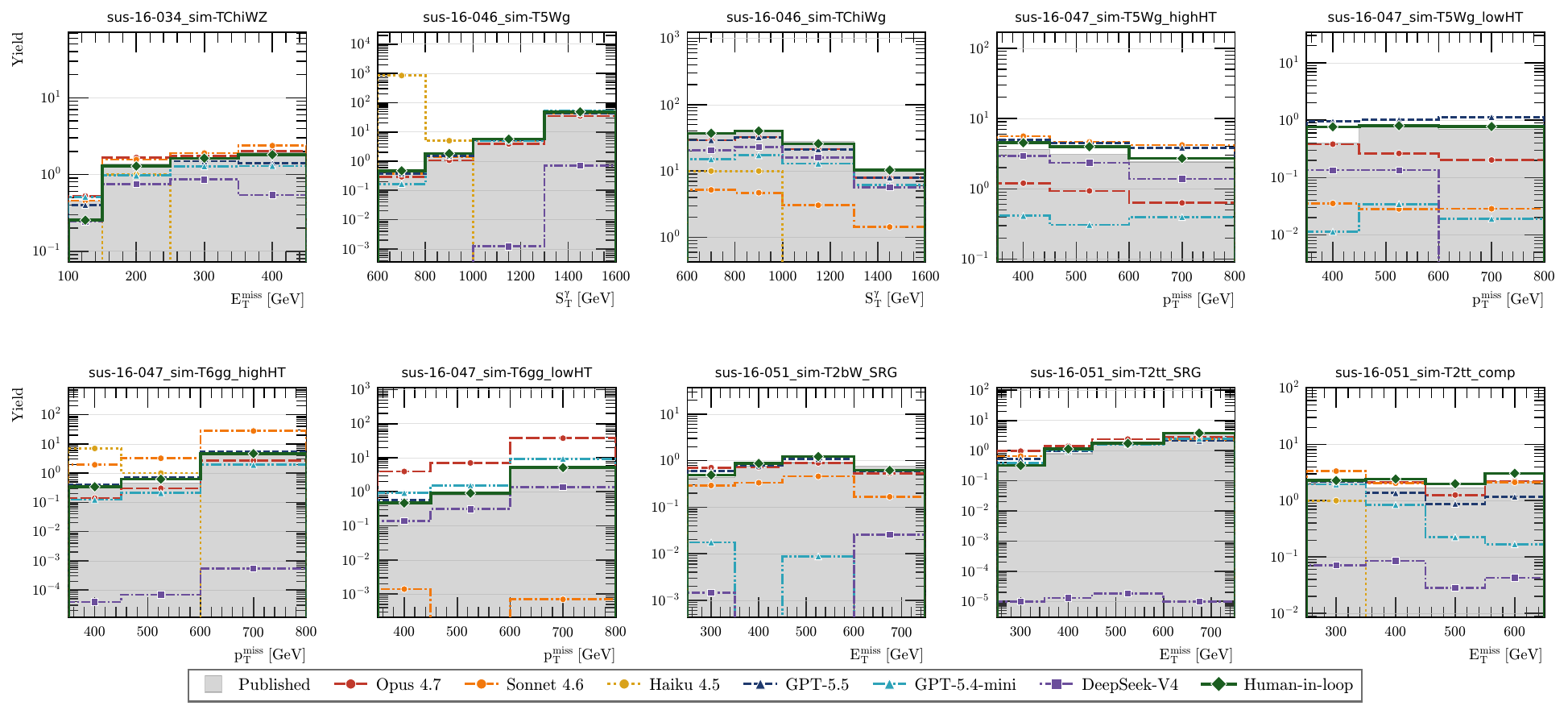}
    \caption{Best-of-runs per agent against the published yields (grey) for
             every \texttt{sim} task in the benchmark. Best run selected according to the relative \(L^2\) error on the absolute binned yields; runs flagged \textsc{fabricated} by the provenance audit are excluded. The dark green diamond-marked line is our
         physicist-in-the-loop reproduction, shown as a
         non-autonomous reference.}
    \label{fig:appendix_sim}
\end{figure}

\begin{figure}
    \centering
    \includegraphics[width=1\linewidth]{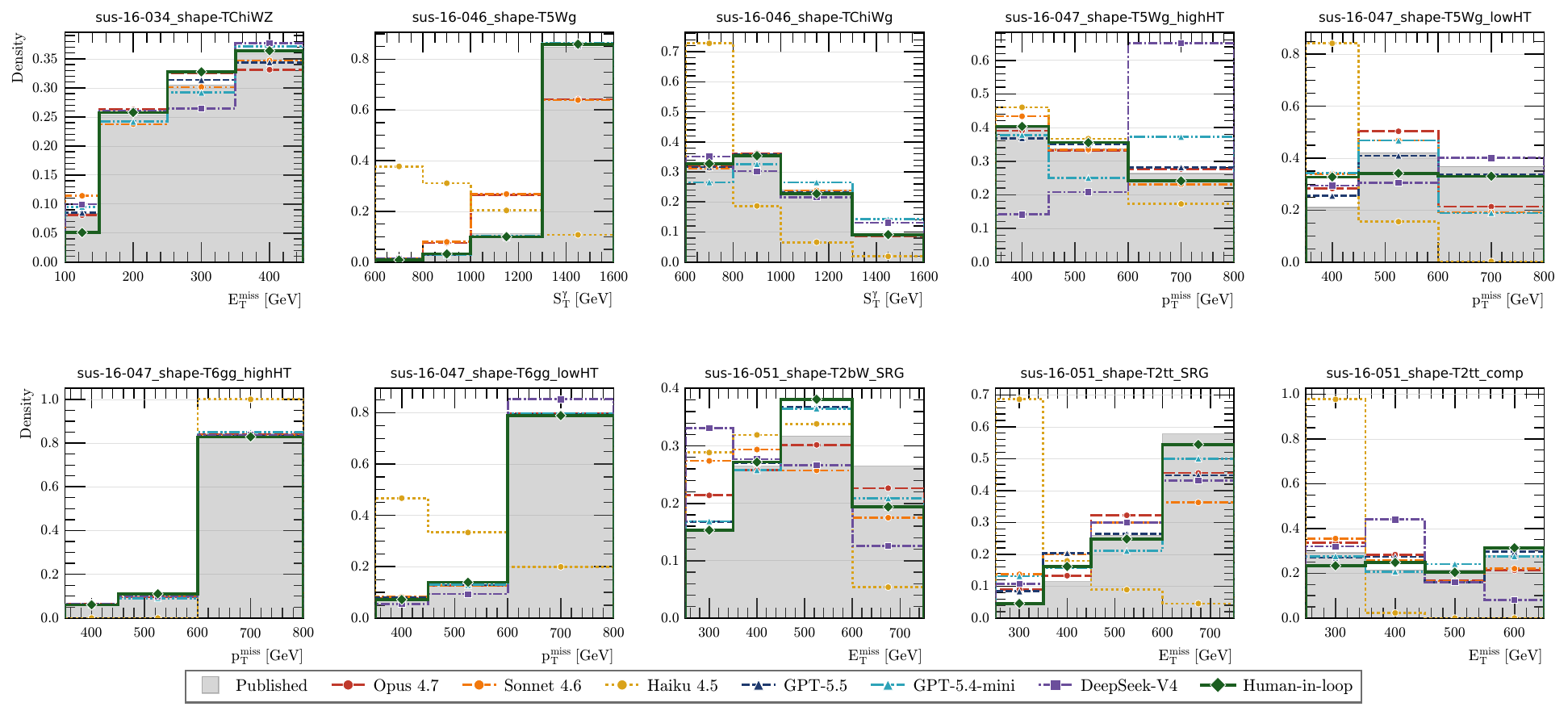}
    \caption{Best-of-runs per agent against the published shape (grey) for
             every \texttt{shape} task; all distributions are unit-normalised.
             Best run selected according to the relative \(L^2\) error on the normalized binned yields; runs flagged
             \textsc{fabricated} by the provenance audit are excluded. The dark green diamond-marked line is our
         physicist-in-the-loop reproduction, shown as a
         non-autonomous reference.}
    \label{fig:appendix_shape}
\end{figure}

\begin{itemize}
    \item Figure~\ref{fig:appendix_pies} shows the fraction of runs by completion status.
    \item Table~\ref{tab:sim_scores} reports relative-\(L^2\) errors on absolute binned yields for the \texttt{sim} tasks.
    \item Table~\ref{tab:sim_shape_scores} reports the corresponding shape-only diagnostic on unit-normalised distributions.
    \item Figure~\ref{fig:pareto_shape} shows model performance on \texttt{shape} tasks and the cost--accuracy Pareto frontier.
    \item Figure~\ref{fig:appendix_sim} shows best-of-runs overlays against the published yields for all \texttt{simulation} tasks.
    \item Figure~\ref{fig:appendix_shape} shows best-of-runs overlays against the published shapes for all \texttt{shape} tasks.
    \item Table \ref{tab:resources-sim} shows the average number of tokens, costs and run-times for each agent.
\end{itemize}

\begin{table}[t]
\centering
\small
\setlength{\tabcolsep}{4pt}
\resizebox{\textwidth}{!}{%
\begin{tabular}{lcccccc}
\toprule
 & \textbf{Claude Code} & \textbf{Claude Code} & \textbf{Claude Code} & \textbf{Codex} & \textbf{Codex} & \textbf{ForgeCode} \\
\textbf{Collider-Bench Tasks} & Opus 4.7 & Sonnet 4.6 & Haiku 4.5 & GPT-5.5 & GPT-5.4-mini & DeepSeek-V4 \\
\midrule
\rowcolor{gray!8}\texttt{sus-16-034\_sim-TChiWZ}      & $\mathbf{0.19{\pm}0.12}$ & $0.30{\pm}0.14$ & 0.893  & $0.29{\pm}0.08$ & $0.65{\pm}0.35$ & $0.80{\pm}0.29$ \\
\texttt{sus-16-046\_sim-T5Wg}                         & $0.51{\pm}0.38$ & $0.61{\pm}0.46$ & 18.0             & $\mathbf{0.13{\pm}0.08}$ & $0.50{\pm}0.45$ & $0.99{\pm}0.01$ \\
\rowcolor{gray!8}\texttt{sus-16-046\_sim-TChiWg}      & $0.49{\pm}0.39$ & $0.93{\pm}0.05$ & $0.92{\pm}0.10$  & $\mathbf{0.28{\pm}0.06}$ & $0.84{\pm}0.23$ & $0.84{\pm}0.27$ \\
\texttt{sus-16-047\_sim-T5Wg\_highHT}                 & $1.43{\pm}1.25$ & $0.69{\pm}0.18$ & $213{\pm}299$    & $\mathbf{0.62{\pm}0.24}$ & $1.72{\pm}0.93$ & $1.41{\pm}1.39$ \\
\rowcolor{gray!8}\texttt{sus-16-047\_sim-T5Wg\_lowHT} & $\mathbf{0.75{\pm}0.17}$ & $1.20{\pm}0.38$ & --            & $0.79{\pm}0.13$ & $1.00{\pm}0.04$ & $0.96{\pm}0.07$ \\
\texttt{sus-16-047\_sim-T6gg\_highHT}                 & $0.64{\pm}0.33$ & $1.27{\pm}0.47$ & $1117{\pm}1918$  & $\mathbf{0.58{\pm}0.37}$ & $1.43{\pm}1.18$ & $157{\pm}255$ \\
\rowcolor{gray!8}\texttt{sus-16-047\_sim-T6gg\_lowHT} & $11.3{\pm}5.0$  & $1.46{\pm}0.80$ & $278{\pm}240$    & $\mathbf{0.45{\pm}0.47}$ & $0.95{\pm}0.05$ & $0.90{\pm}0.16$ \\
\texttt{sus-16-051\_sim-T2bW\_SRG}                    & $\mathbf{0.28{\pm}0.05}$ & $0.82{\pm}0.21$ & --               & $0.71{\pm}0.45$ & $1.41{\pm}0.71$ & $14.9{\pm}24.2$ \\
\rowcolor{gray!8}\texttt{sus-16-051\_sim-T2tt\_SRG}   & $\mathbf{0.42{\pm}0.05}$ & $0.53{\pm}0.21$ & $34.4{\pm}29.9$  & $0.46{\pm}0.04$ & $0.78{\pm}0.37$ & $1.02{\pm}0.03$ \\
\texttt{sus-16-051\_sim-T2tt\_comp}                   & $\mathbf{0.50{\pm}0.40}$ & $0.69{\pm}0.36$ & $125{\pm}210$    & $0.37{\pm}0.04$ & $0.84{\pm}0.17$ & $0.99{\pm}0.01$ \\
\bottomrule
\end{tabular}}%
\caption{Binned yields relative-$L^2$ on the \texttt{sim} tasks
($d=\sqrt{\sum_k(\hat{y}_k-y^\star_k)^2/\sum_k {y^\star_k}^2}$, lower
is better): mean $\pm$ 1$\sigma$ over independent runs; single-run
cells reported as the bare value. Only runs flagged \textsc{fabricated}
by the provenance audit are excluded; all-null and all-zero submissions
score $d{=}1$ (the metric's natural floor for empty output).
Dashed (\texttt{--}) entries indicate that all runs for that
(model, task) cell were flagged as fabricatedß. \textbf{Bold} marks the
best mean per task.}
\label{tab:sim_scores}
\end{table}

\begin{table}[t]
\centering
\small
\setlength{\tabcolsep}{4pt}
\resizebox{\textwidth}{!}{%
\begin{tabular}{lcccccc}
\toprule
 & \textbf{Claude Code} & \textbf{Claude Code} & \textbf{Claude Code} & \textbf{Codex} & \textbf{Codex} & \textbf{ForgeCode} \\
\textbf{Collider-Bench Tasks} & Opus 4.7 & Sonnet 4.6 & Haiku 4.5 & GPT-5.5 & GPT-5.4-mini & DeepSeek-V4 \\
\midrule
\rowcolor{gray!8}\texttt{sus-16-034\_sim-TChiWZ}      & $0.09{\pm}0.03$ & $\mathbf{0.07{\pm}0.02}$ & $1.21{\pm}0.37$  & $0.10{\pm}0.02$ & $0.13{\pm}0.07$ & $0.48{\pm}0.26$ \\
\texttt{sus-16-046\_sim-T5Wg}                         & $0.12{\pm}0.17$ & $0.15{\pm}0.15$ & 1.53             & $\mathbf{0.02{\pm}0.01}$ & $0.25{\pm}0.37$ & $0.19{\pm}0.04$ \\
\rowcolor{gray!8}\texttt{sus-16-046\_sim-TChiWg}      & $0.07{\pm}0.10$ & $0.15{\pm}0.04$ & $1.00{\pm}0.34$  & $\mathbf{0.02{\pm}0.01}$ & $0.22{\pm}0.22$ & $0.45{\pm}0.42$ \\
\texttt{sus-16-047\_sim-T5Wg\_highHT}                 & $\mathbf{0.07{\pm}0.02}$ & $0.29{\pm}0.43$ & $0.90{\pm}0.50$  & $0.08{\pm}0.03$ & $0.17{\pm}0.13$ & $0.30{\pm}0.32$ \\
\rowcolor{gray!8}\texttt{sus-16-047\_sim-T5Wg\_lowHT} & $0.44{\pm}0.26$ & $0.52{\pm}0.42$ & 1.00             & $\mathbf{0.22{\pm}0.06}$ & $0.49{\pm}0.44$ & $0.70{\pm}0.36$ \\
\texttt{sus-16-047\_sim-T6gg\_highHT}                 & $0.38{\pm}0.63$ & $0.51{\pm}0.52$ & $0.79{\pm}0.58$  & $0.37{\pm}0.60$ & $0.36{\pm}0.55$ & $\mathbf{0.18{\pm}0.24}$ \\
\rowcolor{gray!8}\texttt{sus-16-047\_sim-T6gg\_lowHT} & $\mathbf{0.04{\pm}0.03}$ & $0.65{\pm}0.55$ & $0.94{\pm}0.31$  & $0.08{\pm}0.10$ & $0.49{\pm}0.50$ & $0.09{\pm}0.03$ \\
\texttt{sus-16-051\_sim-T2bW\_SRG}                    & $\mathbf{0.27{\pm}0.04}$ & $0.73{\pm}0.75$ & --               & $0.68{\pm}0.44$ & $0.78{\pm}0.61$ & $1.01{\pm}0.56$ \\
\rowcolor{gray!8}\texttt{sus-16-051\_sim-T2tt\_SRG}   & $0.40{\pm}0.04$ & $0.52{\pm}0.24$ & $1.31{\pm}0.31$  & $0.27{\pm}0.05$ & $\mathbf{0.27{\pm}0.09}$ & $1.16{\pm}0.56$ \\
\texttt{sus-16-051\_sim-T2tt\_comp}                   & $\mathbf{0.18{\pm}0.06}$ & $0.26{\pm}0.10$ & $1.06{\pm}0.53$  & $0.35{\pm}0.20$ & $0.87{\pm}0.12$ & $0.38{\pm}0.11$ \\
\bottomrule
\end{tabular}}%
\caption{Shape diagnostic on the \texttt{sim} tasks
($d=\sqrt{\sum_k(\hat{p}_k-p^\star_k)^2/\sum_k {p^\star_k}^2}$ on the
unit-normalised distributions $p_k=\hat y_k/\sum_j \hat y_j$, lower is
better): mean $\pm$ 1$\sigma$ over independent runs of the same
(model, task), reported when $\geq 2$ runs are available; otherwise
the single-run value. Only runs flagged \textsc{fabricated} by the
provenance audit are excluded; all-null and all-zero submissions
score $d{=}1$ (the value the metric assigns to an empty submission).
Dashed (\texttt{--}) entries indicate that all runs for that
(model, task) cell were flagged as fabricated. \textbf{Bold} marks the
best mean per task.}
\label{tab:sim_shape_scores}
\end{table}

\begin{table}[t]
    \centering
    \caption{\ColliderBench{} consumption metrics (mean $\pm 1\sigma$) per agent for the {\tt Simulation} tasks. Tokens are billed totals, cost is the list price for Spring 2026.}
    \vspace{5pt}
    \scriptsize
    \label{tab:resources-sim}
    \begin{tabular}{lcccc}
        \toprule
        Agent & Billed Tokens (M) & Cost (list \$) & Run Time(h) & $n_{\rm tasks}$ \\
        \midrule
        \rowcolor{gray!8}
        Opus 4.7      & $\phantom{0}0.37 \pm \phantom{0}0.11$  & $14.21 \pm 4.56$ & $0.664 \pm 0.180$ & 10 \\
        Sonnet 4.6    & $\phantom{0}0.38 \pm \phantom{0}0.10$  & $\phantom{0}7.63 \pm 1.67$ & $1.095 \pm 0.366$ & 10 \\
        \rowcolor{gray!8}
        Haiku 4.5     & $\phantom{0}0.15 \pm \phantom{0}0.12$  & $\phantom{0}1.09 \pm 0.61$ & $0.299 \pm 0.249$ & 10 \\
        GPT-5.5       & $16.46 \pm \phantom{0}8.53$            & $10.71 \pm 5.04$ & $0.570 \pm 0.217$ & 10 \\
        \rowcolor{gray!8}
        GPT-5.4-mini  & $29.83 \pm 10.43$                       & $\phantom{0}3.66 \pm 1.31$ & $0.864 \pm 0.351$ & 10 \\
        DeepSeek-V4   & $\phantom{0}2.55 \pm \phantom{0}1.71$  & $\phantom{0}0.89 \pm 0.46$ & $1.706 \pm 0.584$ & 10 \\
        \bottomrule
    \end{tabular}
\end{table}

\subsection{Tool Ablation Test}
\label{sec:tool-ablation}

As a tool-availability stress test, we disabled the CLI tool for fast-detector simulation, \texttt{Delphes}, and reran the simulation tasks for GPT-5.5 with the remaining interface unchanged. The agent still received the same papers, prompts, output templates, and analysis environment, but could no longer rely on the standard detector-simulation step. Strong agents recognized that this component of the workflow was unavailable and, without external prompting, implemented simplified parametric detector approximations in Python.

The effect of this ablation is task dependent. For several tasks, the scores with and without \texttt{Delphes} are comparable, indicating that a strong agent can sometimes construct an adequate approximation for the detector-level objects needed by the analysis. However, we found that the CMS-SUS-16-047 tasks degrade more clearly without \texttt{Delphes}. This suggests that the harder tasks are more sensitive to detector-level modeling, such as object reconstruction, missing-momentum response, or bin migration, and cannot be replaced as easily by a lightweight approximation.

We therefore do not interpret this ablation as evidence that detector simulation is generally unnecessary. Rather, it shows that the necessity of a domain tool is itself task dependent. In simpler settings, agents may compensate for missing tools by constructing approximate substitutes; in more detector-sensitive analyses, the validated toolchain remains important. This makes \ColliderBench{} useful not only for testing tool use, but also for testing whether agents can decide when an approximation is adequate and when a missing scientific tool materially changes the result.

\section{LLM Judge}
\label{app:judge}

Every run is post-audited by a separate LLM call, default mode
\texttt{claude-opus-4-6}, held fixed across the study and not
evaluated as an agent). The judge examines the agent's filled
\texttt{results/histogram.yaml}, the hidden reference (used only for
leakage detection), workspace artifacts (\texttt{report.md},
analysis scripts, etc.), and a 50k-char structured extract of the
session log --- not the full trace.

It returns a provenance audit and a short, evidence-based
trajectory narrative (planning, tool use, scientific judgment,
stuck points). The audit can also recover bin values the agent
computed but never wrote into the template, in which case the
benchmark automatically re-scores the run against the recovered
values. The trajectory narrative is recorded but does not enter
the numerical score.

In the aggregate tables we drop only runs the judge invalidated
in full (those we call \textsc{fabricated} in the main text); rescored
runs use the judge's recovered values.

\end{document}